\PassOptionsToPackage{capitalize}{cleveref} 
\documentclass[10pt,twocolumn,letterpaper]{article}

\usepackage{iccv}

\usepackage[ruled,linesnumbered]{algorithm2e}
\usepackage{multirow}
\usepackage{booktabs}
\usepackage{hhline}
\usepackage{tabularx}
\usepackage{bbding}
\usepackage{enumitem}
\usepackage{subcaption}      %
\usepackage{xcolor}
\usepackage{amssymb}
\usepackage{colortbl}
\usepackage{wrapfig}
\usepackage[accsupp]{axessibility}

\definecolor{iccvblue}{rgb}{0.21,0.49,0.74}
\usepackage{gensymb}
\usepackage{amsmath}

\usepackage{bm}
\usepackage{caption}
\usepackage{float}

\usepackage{enumitem}
\usepackage{tikz}
\usepackage{pifont}
\usepackage{bbm}
\usepackage{amsthm,amsmath}
\usepackage{mathrsfs}
\usepackage{mathtools}
\usepackage{color}

\usepackage{balance}

\DeclarePairedDelimiterX{\infdivxbig}[2]{\Big(}{\Big)}{%
  #1\;\delimsize\|\;#2%
}
\newcommand{\infdivbig}{\infdivxbig}

\DeclarePairedDelimiterX{\infdivx}[2]{(}{)}{%
  #1\;\delimsize\|\;#2%
}

\definecolor{lightblue}{rgb}{0.93,0.95,1.0} %
\definecolor{GrayLight}{gray}{0.95}
\definecolor{GrayMedium}{gray}{0.85}
\definecolor{GrayMedium2}{gray}{0.75}
\definecolor{GrayMedium3}{gray}{0.65}
\definecolor{GrayDark}{gray}{0.55}
\newcolumntype{C}{>{\columncolor{lightblue}}c}

\definecolor{iccvblue}{rgb}{0.21,0.49,0.74}
\usepackage[pagebackref,breaklinks,colorlinks,allcolors=iccvblue]{hyperref}

\usepackage{cleveref}        %
\Crefname{equation}{Eq.}{Eqs.}
\Crefname{figure}{Fig.}{Figs.}
\Crefname{section}{Sec.}{Secs.}
\Crefname{table}{Tab.}{Tabs.}
\def\paperID{8533} %
\def\confName{ICCV}
\def\confYear{2025}

\title{Divide-and-Conquer for Enhancing Unlabeled Learning, Stability, and Plasticity in Semi-supervised Continual Learning}

\author{Yue Duan\textsuperscript{\rm 1}\thanks{Equal contribution.}\quad\quad\quad Taicai Chen\textsuperscript{\rm 1}\footnotemark[1] \quad\quad\quad Lei Qi\textsuperscript{\rm 2} \quad\quad\quad Yinghuan Shi\textsuperscript{\rm 1}\thanks{Corresponding author.}  \\
	\textsuperscript{\rm 1}Nanjing University\ \quad\quad\quad \textsuperscript{\rm 2}Southeast University\  \\
	{\tt\small\{yueduan@smail., taicaichen@smail., syh@\}nju.edu.cn,} {\tt\small qilei@seu.edu.cn} \\
}
\begin{document}
\maketitle
\begin{abstract}
Semi-supervised continual learning (SSCL) seeks to leverage both labeled and unlabeled data in a sequential learning setup, aiming to reduce annotation costs while managing continual data arrival. SSCL introduces complex challenges, including ensuring effective unlabeled learning (UL), while balancing memory stability (MS) and learning plasticity (LP).   Previous SSCL efforts have typically focused on isolated aspects of the three, while this work presents USP, a divide-and-conquer framework designed to synergistically enhance these three aspects: (1) Feature Space Reservation (FSR) strategy for LP, which constructs reserved feature locations for future classes by shaping old classes into an equiangular tight frame; (2) Divide-and-Conquer Pseudo-labeling (DCP) approach for UL, which assigns reliable pseudo-labels across both high- and low-confidence unlabeled data; and (3) Class-mean-anchored Unlabeled Distillation (CUD) for MS, which reuses DCP's outputs to anchor unlabeled data to stable class means for distillation to prevent forgetting. Comprehensive evaluations show USP outperforms prior SSCL methods, with gains up to 5.94\% in the last accuracy, validating its effectiveness. The code is available at  \url{https://github.com/NJUyued/USP4SSCL}.
\end{abstract}

\definecolor{ss}{rgb}{1,0.4,0.4}
\definecolor{uu}{rgb}{0,0.6,0}
\definecolor{pp}{rgb}{0,0.4,0.8}

\section{Introduction}
\label{sec:intro}

\begin{figure}
    \centering
    \includegraphics[width=\linewidth]{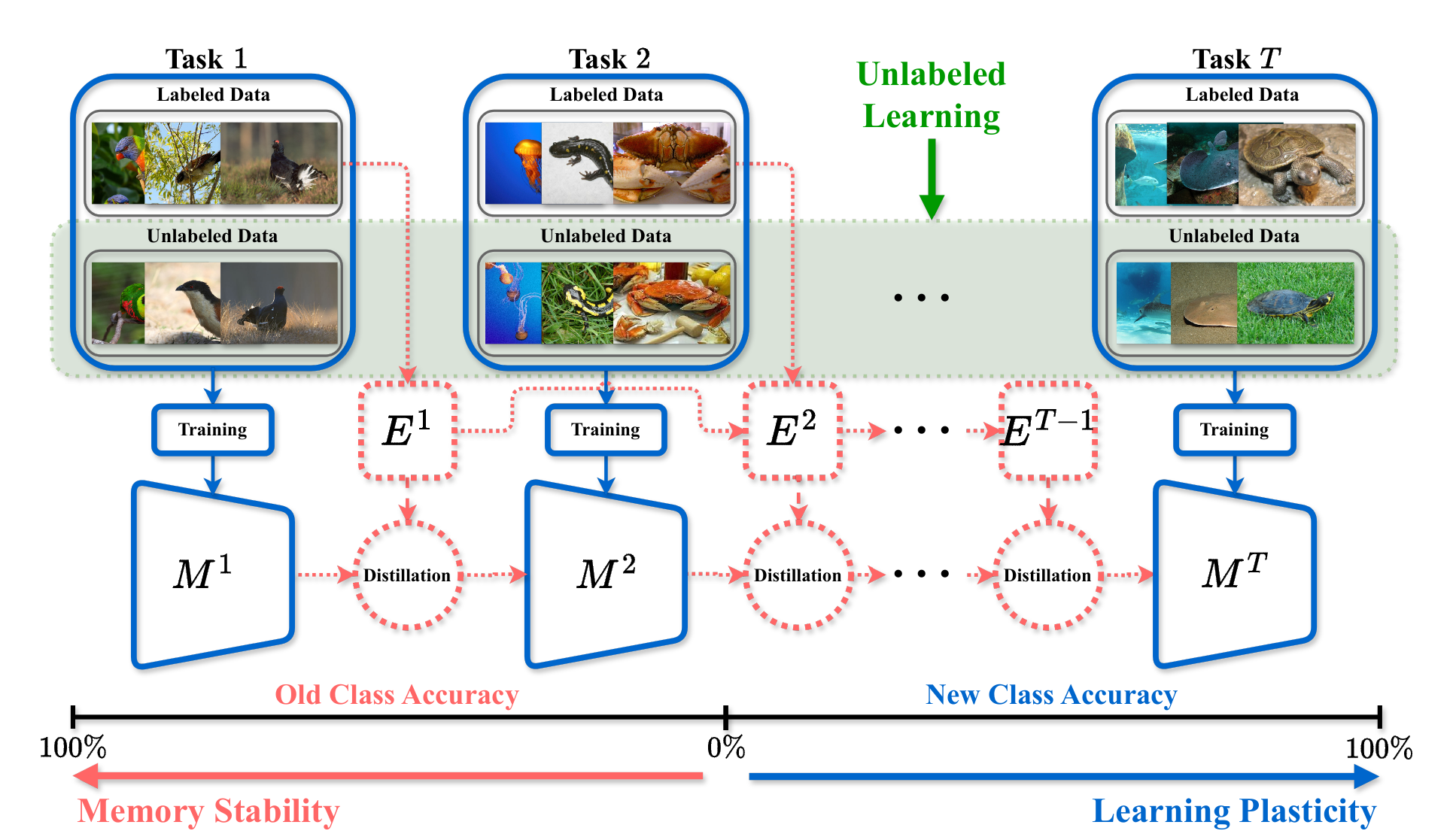}
    \caption{Illustration of the SSCL setting ($E^t$ and $M^t$ indicate the exemplar set and model in task $t$, respectively): in a dynamic data stream containing both labeled and unlabeled data, distinct tasks arrive sequentially with non-overlapping classes across tasks. 
    }
    \label{fig:intro}
\end{figure}

Recently, \textit{continual learning} (CL) has emerged as a promising approach for handling such sequential data arrival scenarios \cite{wickramasinghe2023continual,wang2024comprehensive}. Yet, most existing CL methods rely heavily on fully labeled data, which is often impractical in real-world applications due to high annotation costs, privacy concerns, and limitations in adapting to real-time online scenarios. To address these challenges, researchers have turned to \textit{semi-supervised learning} (SSL) \cite{ouali2020overview,yang2022survey} frameworks, where only a subset of samples requires labeling. Inspired by this paradigm, a new direction called \textit{semi-supervised continual learning} (SSCL) \cite{wang2021ordisco,bagus2022beyond} has emerged, aiming to leverage SSL setups across all tasks within the CL setting, which is illustrated in \Cref{fig:intro}.

SSCL introduces distinct challenges due to the need to continuously learn from both labeled and unlabeled data. This setting demands careful management of the trade-off between \textit{learning plasticity}—the model’s ability to incorporate new knowledge, and \textit{memory stability}—its capacity to retain past knowledge \cite{kong2022balancing,kim2023achieving,zhou2024class}. In SSCL, these dynamics become particularly complex as the model faces the risk of catastrophic forgetting of past tasks, while simultaneously being prone to overfitting on limited labeled samples \cite{cui2023uncertainty,luo2024learning, fan2024dynamic}. Meanwhile, learning from a vast pool of unlabeled data is also challenging, as CL’s anti-forgetting processes can disrupt learning under sparse supervision. Furthermore, standard CL techniques like experience replay face obstacles in SSCL: constrained replay buffers often prioritize labeled samples, limiting the use of valuable information from unlabeled data \cite{smith2021memory, brahma2021hypernetworks, gomes2022survey}.

We argue that addressing SSCL effectively requires a holistic approach that does not overlook any of these three aspects: \textbf{\textcolor{uu}{unlabeled learning (UL)}}, \textbf{\textcolor{ss}{memory stability (MS)}}, and \textbf{\textcolor{pp}{learning plasticity (LP)}}. Although previous approaches have made strides in SSCL, most of them primarily focus on \textit{just one or two of the three core challenges, often leaving the other aspects unaddressed}. For examples, for UL, \cite{smith2021memory} employs pseudo-labeling technique used in SSL to utilize unlabeled data for training and \cite{de2021continual} apply consistency loss to boost the robustness of unlabeled learning; For MS, \cite{fan2024dynamic} proposes dynamic sub-graph distillation (DSGD) to leverages semantic and structural information from unlabeled data, helping the model remain robust to distribution shifts. Motivated by this, we propose a \textit{\textbf{divide-and-conquer approach with three interlinked modules: \textcolor{uu}{\textbf{U}}\textcolor{ss}{\textbf{S}}\textcolor{pp}{\textbf{P}}}, each tailored to enhance one of these aspects while collectively improving the overall SSCL performance.}

The key to building a powerful SSCL lies in designing mechanisms that enable UL-, MS- and LP-components to complement and reinforce each other. With the established  baseline SSCL learner, we first introduce a projection head to produce an additional feature branch, which serves as a unified feature output across all subsequent components aiming to strengthen their coupling for mutual enhancement. \textcolor{pp}{\textbf{(1) For LP}}: We propose a simple yet effective Feature Space Reservation strategy (FSR). Leveraging the equiangular tight frame (ETF) for optimal feature geometry, we obtain anchor vectors that reserve space for future classes. A contrastive-like loss is proposed to align learned data features to these class-specific positions, laying a strong foundation for subsequent CL processes. \textcolor{uu}{\textbf{(2) For UL}}: We propose a Divide-and-Conquer Pseudo-labeling strategy (DCP) to handle high- and low-confidence unlabeled data separately, leveraging two complementary pseudo-labeling techniques. This approach ensures effective utilization of all data while maintaining pseudo-label accuracy, ultimately delivering a robust UL process—\textit{and even offering a ``free lunch" benefit during testing phase}. \textcolor{ss}{\textbf{(3) For MS}}: We repurpose intermediate calculations from DCP to introduce the Class-mean-anchored Unlabeled Distillation (CUD). CUD aggregates the latent relationship between labeled and unlabeled data, enhancing model's resistance to catastrophic forgetting on unlabeled data, thus supporting stable and reliable representation retention in SSCL.

\noindent \textbf{Contribution.} A divide-and-conquer framework, \textcolor{uu}{\textbf{U}}\textcolor{ss}{\textbf{S}}\textcolor{pp}{\textbf{P}}, is proposed that synergistically enhances unlabeled learning (\textcolor{uu}{\textbf{U}}L), memory stability (M\textcolor{ss}{\textbf{S}}), and learning plasticity (L\textcolor{pp}{\textbf{P}}): (1) We introduce a novel pseudo-labeling scheme, which ensures high-quality pseudo-labels across confidence levels, fully utilizing all data to improve UL. (2) We propose a feature space reservation strategy and cross-labeled-unlabeled distillation to jointly enhance LP and MS, helping the model resist forgetting. (3) Extensive evaluations across diverse SSCL settings demonstrate the performance benefits of \textcolor{uu}{\textbf{U}}\textcolor{ss}{\textbf{S}}\textcolor{pp}{\textbf{P}}, with up to a 4.10\% gain in average accuracy.

\section{Related Works}
\label{sec:rw}

\subsection{Continual Learning}
Continual learning (CL) addresses catastrophic forgetting during incremental learning. As summarized in \cite{zhou2024class}, mainstream approaches fall into three categories: \textit{replay}-based, \textit{knowledge distillation}-based, and \textit{dynamic network}-based methods. Replay-based methods retain/rehearse past data through stored exemplars \cite{bang2021rainbow, zhao2021memory, de2021continual} or synthetic generation \cite{petit2023fetril, jodelet2023class, gao2023ddgr}. Knowledge distillation-based methods transfer knowledge from old to new models through distillation of logits \cite{rebuffi2017icarl, zhang2020class,smith2021always}, features \cite{lu2022augmented, park2021class, kang2022class}, or relations \cite{asadi2023prototype, liu2022model}. Dynamic network-based methods expand architectures via neuron \cite{li2019learn}, backbone \cite{yan2021dynamically, wang2022foster}, or prompt \cite{wang2022learning, smith2023coda} growth.

While CL methods are often categorized distinctly, their boundaries are fluid. Replay and distillation synergize as core anti-forgetting mechanisms across paradigms. Our method similarly harnesses their complementary strengths to combat catastrophic forgetting.

\subsection{Semi-supervised Learning}
\label{sec:ssl}
Semi-supervised learning (SSL) aims to reduces the dependency of deep learning models on labeled data by leveraging abundant unlabeled data. Early SSL methods can be broadly categorized into \textit{pseudo-labeling} and \textit{consistency regularization} methods. Pseudo-labeling methods expand the training set by assigning predictions on unlabeled data as pseudo-labels \cite{lee2013pseudo,xie2020self,duan2024roll,chen2024pg}. In contrast, consistency regularization methods enforce similar predictions across augmented versions of input samples, enhancing generalization boundaries through teacher-student interactions \cite{tarvainen2017mean} or by applying diverse perturbations to inputs \cite{miyato2018virtual,sohn2020fixmatch,duan2024mutexmatch}. FixMatch \cite{sohn2020fixmatch} introduced a simple yet effective framework that integrates these two strategy: it applies weak and strong augmentations to unlabeled data and uses high-confidence predictions on weakly-augmented samples as pseudo-labels for the strongly-augmented counterparts, enforcing a strong-weak consistency regularization on unlabeled data. This approach demonstrated remarkable performance and has become a foundational benchmark in SSL, inspiring many subsequent methods that adjust pseudo-labeling strategies for unlabeled data \cite{ zheng2022simmatch, zheng2023simmatchv2,duan2023towards} or modify confidence thresholding schemes \cite{wang2022freematch, chen2023softmatch, chen2023boosting}.

While FixMatch and similar methods have achieved significant success in SSL, current SSL methods still fall short in addressing learning scenarios where data distribution or class compositions may shift over time.

\subsection{Semi-supervised Continual Learning}
Existing continual learning methods generally rely on a fully supervised setup, whereas semi-supervised continual learning (SSCL) more realistically assumes that only a limited number of samples are labeled at each task. 
The core challenge in SSCL is effectively utilizing unlabeled data to mitigate catastrophic forgetting. CNNL \cite{boschini2022continual} fine-tunes its incremental learner by generating pseudo-labels for the unlabeled data, enabling self-training. DistillMatch \cite{smith2021memory} employs knowledge distillation with prediction consistency on unlabeled data. It also optimizes an out-of-distribution detector to identify task-specific representations. Pseudo-gradient learner \cite{luo2024learning} introduces a gradient predictor using labeled data to estimate gradients for unlabeled data, thereby avoiding the potential risks of pseudo-labeling. ORDisCo \cite{wang2021ordisco} learns continuously from partially labeled data using a classifier-equipped conditional GAN and performs online data replay. MCSSL \cite{brahma2021hypernetworks} extends ORDisCo into a meta-learning framework. DSGD \cite{fan2024dynamic} introduces a dynamic subgraph distillation method that leverages semantic and structural information for more stable knowledge distillation on unlabeled data. %

Unlike previous methods that focus on individual aspects of UL, MS, or LP, our proposed method integrates all three into a unified framework, aiming for a synergistic effect that amplifies their combined impact.

\section{Methods}

\begin{figure}
    \centering
    \includegraphics[width=0.99\linewidth]{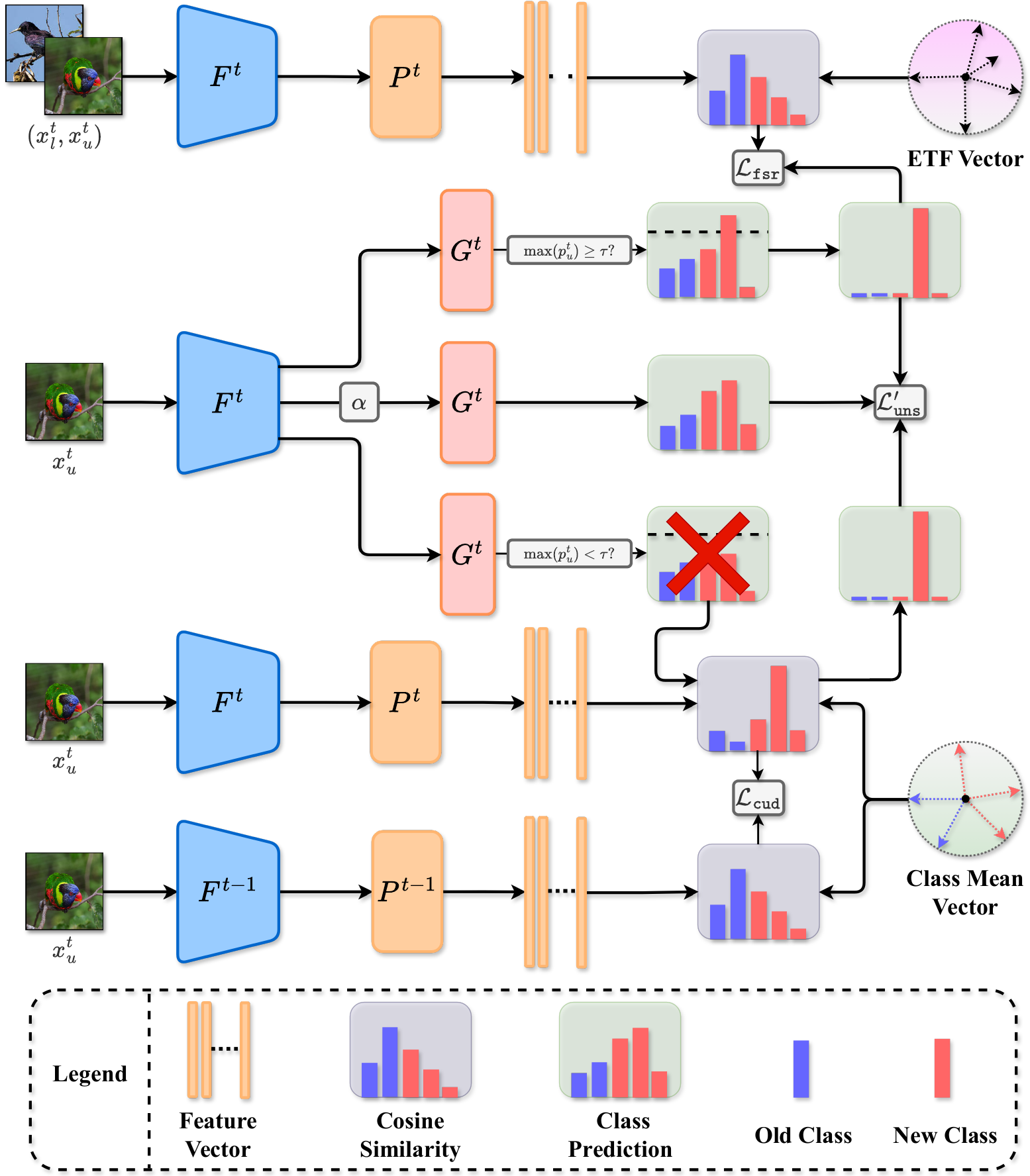}
    \caption{\textbf{Overview of USP.} All losses of USP uniformly utilize the features output by a projection head $P^t$ to enhance synergy through coupling. (1) $\mathcal{L}_{\mathtt{fsr}}$ for LP (\cref{sec:etf}): Given the labeled and unlabeled data $(x^t_l,x^t_u)$, a contrastive loss aligns their features with pre-computed ETF vectors that represent the optimal geometric structure for classification, preventing feature space conflicts between new and old classes during learning; (2) $\mathcal{L}'_{\mathtt{uns}}$ for UL (\cref{sec:dcp}): For $x^t_u$, we compute its class prediction $p^t_u$ and employ a confidence-based divide-and-conquer approach, which leverages complementary pseudo-labels based on classifier and NCM to provide a robust UL process; (3) $\mathcal{L}_{\mathtt{cud}}$ for MS (\cref{sec:cud}): Reusing intermediate results from DCP, we anchor the features of unlabeled data to class mean vectors enriched with information of labeled data, mitigating catastrophic forgetting in SSCL.}
    \label{fig:met}
\end{figure}

\subsection{Baseline SSCL Learner}
\label{sec:bsl}
Denoting the input space as $\mathcal{X}$ and the label space as $\mathcal{Y} = \{1, ..., K\}$ over $K$ classes, we formally define the problem of continual learning (CL) as follows: Given training data that arrives sequentially as a sequence of $T$ tasks, each task $t$ is associated with a dataset $D^t\subseteq \mathcal{X}^t\times\mathcal{Y}^t$, where $t \in \{1, \dots, T\} $ and $\mathcal{Y} ^1 \cap \mathcal{Y} ^2 \cap \cdots  \cap \mathcal{Y} ^T = \emptyset $. Same below, the superscript of $x^t$ is always used to indicate the variable $x$ at different task  $t$. The learning process is conducted task-by-task, and during the training of task $t$, only the current dataset $D^t $ is accessible, while data from previous tasks is systematically discarded. Note that in actual continuous learning settings, a relatively small memory buffer is usually reserved to store past examples to help the model alleviate catastrophic forgetting, which is denoted as $E^t$. $E^t=E^{t,(1)}\cup \cdots \cup E^{t,(K)}$ and $E^{t,(i)}$ is the exemplar set of old class $i$ containing exemplars stored from the datasets in previous $t-1$ tasks.

In semi-supervised continual learning (SSCL), each dataset $D^t $ is partially labeled and can be divided into two subsets: $D^t = D^t_l \cup D^t_u$, where $D^t_l \subseteq \mathcal{X}^t\times\mathcal{Y}^t$ denotes the labeled subset and $D^t_u \subseteq \mathcal{X}^t $ denotes the unlabeled subset, with $|D^t_l| \ll |D^t_u | $. In breif,  we can review
SSCL as a optimization task on the model parameterized by $\theta$:
\begin{equation}
  \mathop{\min}_{\theta}  \sum^{T}_{t=1} \mathcal{L}_{\mathtt{ssl}}(D^t)+\mathcal{L}_{\mathtt{cl}}(E^t), 
  \label{eq:obj}
\end{equation}
where 
$\mathcal{L}_{\mathtt{ssl}}$ is the semi-supervised learning (SSL) loss and $\mathcal{L}_{\mathtt{cl}}$ is the CL
loss. 

We propose a divide and conquer approach to \textbf{U}nlabeled learning, \textbf{S}tability, and \textbf{P}lasticity (\textbf{USP}), which is shown in \Cref{fig:met}.  Overall, we first set a feature extractor $F(\cdot)$, a classifier $G(\cdot)$ and a projection head $P(\cdot)$. Following \cite{fan2024dynamic}, we build a basic SSCL learner using FixMatch \cite{sohn2020fixmatch} for $\mathcal{L}_{\mathtt{ssl}}$, and adopt iCaRL \cite{rebuffi2017icarl} or DER \cite{yan2021dynamically} for $\mathcal{L}_{\mathtt{cl}}$ (DER's details are deferred to Sec. \ref{apx:axpintroduce_der} of supplementary).

\textbf{(1)} $\mathcal{L}_{\mathtt{ssl}}$. Given the labeled data $\{x^{t}_{l},y^{t}_{l}\}\subset D^{t}_l$ and the unlabeled data  $x^{t}_{u}\in D^{t}_u$,  following the most popular pseudo-labeling based SSL method FixMatch, the terms $\mathcal{L}_{\mathtt{ssl}}$ in \Cref{eq:obj} can be decomposed into two
loss terms: $\mathcal{L}_{\mathtt{ssl}}(D^t)= \mathcal{L}_{\mathtt{sup}}(D^t_{l}) + \lambda_{\mathtt{uns}}\mathcal{L}_{\mathtt{uns}}(D^t_{u}) $, where $\lambda_{\mathtt{uns}}$ is the loss weight.  Denoting the model prediction of $x$ as $p^t_{x}=G^t(F^t(x))$, we define the supervised loss as 
\begin{equation}
    \mathcal{L}_{\mathtt{sup}}(D^t_{l}) = \mathbb{E}_{(x^t_l,y^t_l)\sim D^t_l} \left[ H\left(p^t_{x^t_{l}}, y^t_{l}\right) \right],
\end{equation}
where $ H(\cdot, \cdot) $ is the standard cross-entropy loss. Denoting $\Tilde{x}=\max(x)$ and $\hat{x}=\arg\max(x)$, the unsupervised loss is defined as
\begin{equation}
\mathcal{L}_{\mathtt{uns}}(D^t_{u}) = \mathbb{E}_{x^t_u\sim D^t_u} \left[ \mathbbm{1}\Big(\Tilde{p}^t_{x^t_{u}} \geq \tau\Big) H\Big(p^t_{\alpha(x^t_{u})}, \hat{p}^t_{x^t_{u}}\Big)\right],
\label{eq:unsup}
\end{equation}
where $\mathbbm{1}(\cdot)$ is the indicator function, $\hat{p}^t_{x^t_{u}}$ is the hard pseudo-label of $p^t_{x^t_{u}}$ and  $\alpha(\cdot)$ represents a strong data augmentation function\footnote{This implies that we employ \textit{consistency regularization} to enhance the model's prediction stability for semantically similar images. More details of this technique can be found in \Cref{sec:ssl}.} and $\tau$ is a confidence threshold for $\Tilde{p}^t_{x^t_{u}}$ \textcolor{black}{(\ie, the maximum softmax probability)} to select pseudo-labels that are more likely to be correct.

\textbf{(2)} $\mathcal{L}_{\mathtt{cl}}$. With the exemplar management of iCaRL \cite{rebuffi2017icarl} (see Sec. \ref{apx:exemplar_management} of supplementary for details), given $x^{t}_e \in E^t$, we utilize the knowledge distillation loss for $\mathcal{L}_{\mathtt{cl}}$ in \Cref{eq:obj} by encouraging the current model to output the same prediction $p^t_e=G^t(F^t(x^t_{e}))$ as that of the old model: %
\begin{equation}
\mathcal{L}_{\mathtt{cl}}(E^t)= \mathbb{E}_{x^t_e\sim E^t} \left[\mathtt{KL}\infdivbig{\frac{p^t_e}{\beta}}{\frac{p^{t-1}_e}{\beta}}\right],
\label{eq:cl}
\end{equation}
where $\mathtt{KL}(\cdot\parallel \cdot)$ is the KL-divergence and $\beta$ is the temperature parameter. 

Note that from here on, all feature vectors extracted from an input $x$ mentioned in the following text are denoted as $f^t_x$, which are output by the projection head $P(\cdot)$ and normalized using $L_2-$normalization, \ie, $f^t_x=\frac{P^t(F^t(x))}{\|P^t(F^t(x))\|_{2}}$ (and after any operation on the features, they will be renormalized). All components of USP consistently leverage the output features from $P(\cdot)$, \textit{aiming to strengthen the coupling of all components for mutual reinforcement.}

\subsection{ETF-Based Feature Space Reservation}
\label{sec:etf}
We begin by enhancing the plasticity of USP. Inspired by \cite{pernici2021class,zhou2022forward,yang2022neural}, we aim to design a \textbf{F}eature \textbf{S}pace \textbf{R}eservation (\textbf{FSR}) method for accommodating upcoming new classes without interfering with the feature patterns retained for previously learned classes. 

We aim to utilize a simple yet effective contrastive learning loss to align sample features of each class with a set of predefined class prototype features derived from a simplex equiangular tight frame (ETF) for the entire label space. \textcolor{black}{The ETF is inspired by neural collapse phenomenon (more details and the calculation of ETF can be found in Sec. \ref{apx:etf_matrix} of supplementary), which indicates that the final-layer features of samples within the same class collapse to a single vertex. And the vertices of all classes align with class prototypes that form an ETF, which refers to a matrix $\mathcal{E}\in\mathbb{R}^{d\times k}$ ($d$ is a predefined parameter, $k$ is the total number of classes). Crucially, the ETF structure pre-defines maximally separated prototype positions in the feature space, inherently reserving geometric capacity for future unseen classes while preserving semantic discrimination of learned ones.} Each column vector $\mathcal{E}_{:,i} \in \mathbb{R}^{d}$ in ETF can be considered as the corresponding prototype for class $i$. As the ETF corresponds to the optimal geometric structure for classification, anchoring the continually arriving class features to the different ETF-prototype-vectors as learning targets fundamentally enhances the plasticity of SSCL models.  With obtained $\mathcal{E}_{:,i}$, denoting the cosine similarity as $S(\cdot,\cdot)$, we define the following contrastive loss for feature alignment on both the labeled and unlabeled data:
\begin{align}
& \mathcal{L}_{\mathtt{fsr}}(D^t) = \mathbb{E}_{(x^t_l,y^t_l)\sim D^t_l} \left[ -\mathrm{log}\,\frac{\exp\Big(\frac{S(\mathcal{E}_{:,y^t_l},f^{t}_{x^t_{l}})}{\gamma}\Big)}{\sum_{i=1}^{k}\exp\Big(\frac{S(\mathcal{E}_{:,i},f^{t}_{x^t_{l}})}{\gamma}\Big)}\right] \nonumber \\
   & + \mathbb{E}_{x^t_u\sim D^t_u} \left[\mathbbm{1}\Big(\Tilde{p}^t_{x^t_{u}} \geq \tau\Big)   -\mathrm{log}\,\frac{\exp\Big(\frac{S(\mathcal{E}_{:,\hat{p}^t_{x^t_{u}}},f^{t}_{x^t_{u}}) }{\gamma}\Big)}{\sum_{i=1}^{k}\exp\Big(\frac{S(\mathcal{E}_{:,i},f^{t}_{x^t_{u}})}{\gamma}\Big)}\right],
\end{align}
where $\gamma$ is the temperature parameter.
We still use the confidence threshold to filter the pseudo-labels on unlabeled data to ensure the robustness of FSR as much as possible.

\subsection{Divide and Conquer Pseudo-Labeling}
\label{sec:dcp}
For unlabeled learning, considering that we adopt a pseudo-labeling framework similar to FixMatch for $\mathcal{L}_{\mathtt{uns}}$, ensuring the accuracy of pseudo-labels is paramount. Thus, we propose  \textbf{D}ivide-and-\textbf{C}onquer \textbf{P}seudo-labeling (\textbf{DCP}) that combines the two complementary classification methods to leverage their respective strengths in SSCL for $\mathcal{L}_{\mathtt{uns}}$.
\subsubsection{Training Phase}
\label{sec:tp}
In FixMatch and its adaptation as a basic learner in \cite{fan2024dynamic} (\Cref{eq:unsup}), pseudo-labeling involves assigning hard labels by applying a  threshold to the softmax outputs of classifier logits on unlabeled data. However, this will result in low-confidence samples not participating in training and resulting in a waste of data. This is a drawback of classifier-classification, because the pseudo-labels of low-confidence samples are likely to be wrong and harm training. 
Thus, we consider another nearest class mean (NCM) classification method in CL \cite{mensink2012metric}, which classifies samples by matching them to prototype vectors—incrementally computed as the average features of observed examples, for pseudo-labeling on samples with low confidence.

\begin{figure}[t]
    \centering
    \includegraphics[width=\linewidth]{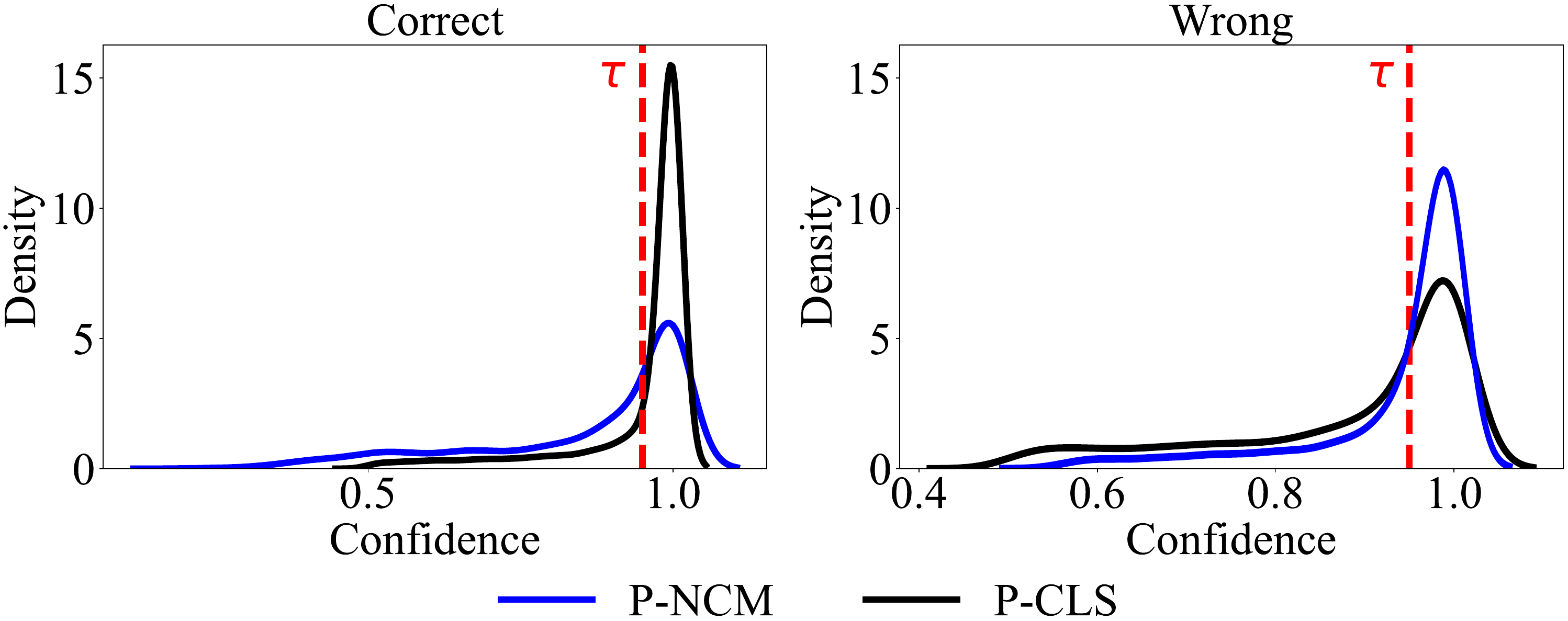}
    \caption{Kernel density estimation (KDE) of confidence distributions of pseudo-labels on 5-task CIFAR10-30 (see \cref{sec:set} for the experimental setting). We show the confidence of classifier-based pseudo-labels (``P-CLS''), divided into correct and incorrect parts, and the re-partitioning of these corresponding examples if the pseudo-labels are providing by NCM (``P-NCM''). The red dash lines indicate the confidence threshold $\tau$ in \Cref{eq:unsup}.
}
    \label{fig:dcp}
\end{figure}

The work in \cite{cui2023uncertainty} argues that NCM-like classification could maintain more stable performance in CL by measuring distances between test image features and class prototypes, which depend only on the parameters of the backbone model. In contrast, classifier-based predictions require feature input into an expanding fully connected layer, which is updated only for the current session, often leading to classification instability in incremental tasks. In this work, as illustrated in \Cref{fig:dcp}, we observe that classifier-based pseudo-labeling can yield highly reliable results at high-confidence levels but lack reliability for low-confidence predictions. NCM classification, on the other hand, maintains a relatively more robust accuracy across all confidence level of classifier predictions. Given discussed above, for high-confidence predictions, DCP apply hard pseudo-labels from the classifier outputs, while low-confidence predictions are assigned pseudo-labels using NCM classification. 

In our NCM classification, considering that the ground-truth of unlabeled data is unknown, the prototype vector is computed for each observed class by averaging the feature vectors of all exemplars selected from labeled data in that class. First, we separate the labeled data by class, where $C^{t,(i)}$ represents the labeled set of class $i$. Then, we compute the feature mean of class $i$ by $\mu_{C^{t,(i)}}=\frac{\sum_{c\in C^{t,(i)}}f^t_c}{|C^{t,(i)}|}$. Finally, the class label $q_{x^t_u}$ for an unlabeled sample $x^t_u$ is assigned based on $\mu_{C^{t,(i)}}$ with the highest similarity to the input’s features, \ie,

\begin{equation}
q_{x^t_u}=\mathop{\arg\max}_{i=1,\cdots,|\mathcal{Y}^t|}S\Big(f^t_{x^t_u},  \mu_{C^{t,(i)}}\Big).
\end{equation}

Hereafter, we can rewrite \Cref{eq:unsup} as
\begin{align}
    \mathcal{L}_{\mathtt{uns}}(D^t_{u})'& = \mathcal{L}_{\mathtt{uns}}(D^t_{u}) +  \nonumber \\
    &
    \mathbb{E}_{x^t_u\sim D^t_u} \left[\mathbbm{1}\Big(\Tilde{p}^t_{x^t_u} < \tau \Big) H\Big(p^t_{\alpha(x^t_{u})}, q_{x^t_u}\Big)\right]. 
\end{align}

\subsubsection{Testing Phase}
We can also treat the testing phase as a pseudo-labeling process, leveraging the benefits of DCP in this context. The key difference is that, for NCM classification of low-confidence samples, we rely on the full exemplar set $E^{t}$ rather than the labeled dataset of a specific task. Given a test sample $u^t$, the predicted label $q_{u^t}$ is assigned by

\begin{equation}
    q_{u^t} =
    \begin{cases}
        \hat{p}^t_{u^t}, & \text{ $\Tilde{p}^t_{u^t} \ge \tau$} \\
        \begin{aligned}
            \mathop{\arg\max}_{i=1,\cdots,k} S\Big(f^t_{u^t} , \mu_{E^{t,(i)}} \Big)
        \end{aligned}, & \text{ $\Tilde{p}^t_{u^t} < \tau$}
    \end{cases},
\end{equation}
where $k$ is the number of class observed so far.

\subsection{Class-Mean-Anchored Unlabeled Distillation}
\label{sec:cud}
Finally, we enhance the stability of USP from a distillation perspective. Knowledge distillation (KD) is a common approach in CL that uses frozen models or stored features and probabilities from prior tasks as a ``teacher'' to guide the active ``student'' model on the new task. In traditional CL, KD typically focuses on independently distilling information from exemplars (\eg, \Cref{eq:cl}), often focusing on labeled samples alone. However, this is insufficient in the SSCL setting, where the forgetting of unlabeled data contributes significantly to catastrophic forgetting. Thus, we introduce \textbf{C}lass-mean-anchored \textbf{U}nlabeled \textbf{D}istillation (\textbf{CUD}), which efficiently reuses the class mean features of labeled data computed in \cref{sec:tp}. Denoting the class mean feature matrix as $M = \begin{bmatrix} \mu_{C^{t,(1)}} \\ \vdots \\ \mu_{C^{t,(i)}} \end{bmatrix}$, we define CUD loss as
\begin{equation}
\mathcal{L}_{\mathtt{cud}}(D^t_u)= \mathbb{E}_{x^t_u\sim D^t_u} \left[\mathtt{KL}\infdivbig{\frac{S(f^{t}_{x^t_{u}},M)}{\xi}}{\frac{S(f^{t-1}_{x^t_{u}},M)}{\xi}}\right]
    ,
\label{eq:ukd}
\end{equation}
where $\xi$ is the temperature parameter. CUD distills the combined relationships between labeled and unlabeled data by anchoring unlabeled samples to the stable class mean features derived from labeled data. This  encourages the model to develop more reliable and robust representations, effectively enhancing the stability of USP.

Finally, the total loss of USP can be presented as
\begin{equation}
    \mathcal{L}=\mathcal{L}_{\mathtt{sup}}+\lambda_{\mathtt{uns}}\mathcal{L}'_{\mathtt{uns}}+\lambda_{\mathtt{cl}}\mathcal{L}_{\mathtt{cl}}+\lambda_{\mathtt{fsr}}\mathcal{L}_{\mathtt{fsr}}+\lambda_{\mathtt{cud}}\mathcal{L}_{\mathtt{cud}}.
\end{equation}

\newcommand{\resb}[3]{#1$\pm$#2~(\textcolor{blue}{$\uparrow$#3})}
\newcommand{\resr}[3]{#1$\pm$#2~(\textcolor{red}{$\downarrow$#3})}
\newcommand{\tol}[3]{#1\ensuremath{_{\pm\mathtt{#2}}^{\textcolor{blue}{\uparrow\mathtt{#3}}}}}
\newcommand{\toll}[3]{#1\ensuremath{_{\pm\mathtt{#2}}^{\textcolor{red}{\downarrow\mathtt{#3}}}}}
\newcommand{\myboxc}[1]{\tikz[baseline=(MeNode.base)]{\node[ fill=red!15](MeNode){#1};}}
\newcommand{\myboxb}[1]{\tikz[baseline=(MeNode.base)]{\node[ fill=blue!25](MeNode){#1};}}
\newcommand{\mystd}[1]{$\pm$#1}

\definecolor{lightblue}{rgb}{0.855, 0.91, 0.988}
\definecolor{GrayMedium}{gray}{0.85}

\begin{table*}[ht]
\centering
\caption{Average and last accuracy on 5-task CIFAR10-$X$ and 10-task CIFAR100-$X$ with $X$ labels per class. 
We provide comparisons with multiple baseline methods reported in DSGD \cite{fan2024dynamic}, which use the same baseline SSCL learner as ours. 
We mark out \textbf{the best result}. }
\scriptsize
\setlength{\tabcolsep}{3.0mm}{
\begin{tabular}{c|cc|cc|cc|cc|cc|cc}
\toprule
\multirow{2}{*}{\textbf{Method}} & \multicolumn{2}{c|}{\textbf{CIFAR10-30}} & \multicolumn{2}{c|}{\textbf{CIFAR10-150}} & \multicolumn{2}{c|}{\textbf{CIFAR100-20}} & \multicolumn{2}{c|}{\textbf{CIFAR100-25}}  & \multicolumn{2}{c|}{\textbf{CIFAR100-80}} & \multicolumn{2}{c}{\textbf{CIFAR100-125}} \\
 & \textbf{Avg} & \textbf{Last} & \textbf{Avg} & \textbf{Last} & \textbf{Avg} & \textbf{Last} & \textbf{Avg} & \textbf{Last} & \textbf{Avg} & \textbf{Last} & \textbf{Avg }& \textbf{Last} \\
\midrule
iCaRL \cite{rebuffi2017icarl} & 34.16 & 21.84 & 60.86 & 53.65 & 26.43 & 13.92 & 28.14 & 15.29 & 36.32 & 19.10 & 44.14 & 30.73 \\
DER \cite{yan2021dynamically} & 40.41 & 31.48 & 64.77 & 61.06 & 31.01 & 23.53 & 32.82 & 26.53 & 53.32 & 41.55 & 57.21 & 48.86 \\
CCIC \cite{boschini2022continual} & - & 55.20 & - & 74.30  & - & - & - & 29.50 & - & - & - & 44.30 \\
ORDisCo \cite{wang2021ordisco} & - & - & 74.77 & 65.91 & - & - & - & - & - & - & - & -\\
NNCSL \cite{kang2023soft} & - & - & - & - & 55.19 & 43.53 & 57.45 & 46.00 & 67.27 & 55.35 & 67.58 & 56.40 \\
\midrule
iCaRL\&Fix \cite{fan2024dynamic} & 45.98 & 30.71 & 78.36 & 69.08 & 45.75 & 23.40 & 49.83 & 31.25 & 53.46 & 32.21 & 56.87 & 41.38 \\
+ DSGD \cite{fan2024dynamic} & 77.33 & \textbf{76.41} & 84.14 & \textbf{79.69} & 52.80 & 35.47 & 53.42 & 35.95 & {57.92} & 37.81 & 58.08 & {43.14} \\
\rowcolor{GrayMedium} \textbf{+ \textcolor{uu}{\textbf{U}}\textcolor{ss}{\textbf{S}}\textcolor{pp}{\textbf{P}} (Ours)} & 79.66 & 70.43 & \textbf{84.78} & 78.21 & 53.20 & 41.30 & 54.36 & 38.25 & 58.59 & 44.20 & 59.96 & 43.80\\
\midrule
DER\&Fix \cite{fan2024dynamic} & 66.71 & 61.41 & 81.10 & 77.00 & 51.76 & 40.86 & 52.03 & 44.47 & 64.03 & 50.25	& 66.69 & 53.57	\\
+ DSGD \cite{fan2024dynamic} & 75.04 & 72.59 & 83.08 & 79.39 & 55.63 & 44.63 & 57.94 & 46.68 & 65.48 & 55.40 & 69.14 & 58.50 \\
\rowcolor{GrayMedium} \textbf{+ \textcolor{uu}{\textbf{U}}\textcolor{ss}{\textbf{S}}\textcolor{pp}{\textbf{P}} (Ours)} & \textbf{81.43} & 73.65 & 84.43 & 77.74 & \textbf{58.79} & \textbf{45.22} & \textbf{59.87} & \textbf{47.44} & \textbf{68.67} & 
\textbf{60.45} & \textbf{71.60} & \textbf{63.08} \\
\bottomrule
\end{tabular}
}
\label{tab:expr_result_cifar}
\end{table*}

\section{Experiments}
\label{sec: exper}

\subsection{Experimental setting}
\label{sec:set}
\textbf{Dataset.} 
We conduct comparative experiments on CIFAR-10, CIFAR-100, and ImageNet-100 to evaluate our method. CIFAR-10 \cite{krizhevsky2009learning} is a dataset with 10 classes, containing 50,000 training images and 10,000 test images, with each image sized 32 × 32. CIFAR-100 \cite{krizhevsky2009learning} is similar to CIFAR-10, but it contains 100 classes, with each class having 500 training images and 100 test images. ImageNet-100 \cite{tian2020contrastive} is a 100-class subset of the ImageNet-1k,
with each class containing 1,300 training images and 500 test images. Additionally, we evaluate on a more challenging few-shot SSCL task on the 
CUB \cite{chaudhry2018efficient} dataset, consisting of 200 bird species with 6,000 training images and 6,000 test images.

\noindent\textbf{Task Settings.} For CIFAR-10, CIFAR-100 and ImageNet-100 datasets, we follow DSGD \cite{fan2024dynamic}, where we sequentially train all 10, 100 and 100 classes in increments of 2, 10 and 10 classes per task, respectively. For CIFAR-10, we use two levels of supervision, with 30 and 150 labeled images per class. For CIFAR-100, we use four levels of supervision, with 20, 25, 80 and 125 labeled images per class.  For ImageNet-100, we mainly use two levels of supervision, with 13 and 100 labeled images per class. To simplify notation, we denote the benchmark as \textit{Dataset-X (number of labeled samples per class)}. For example, CIFAR10-30 indicates CIFAR-10 with 30 labeled samples per class. For CUB, we follow UaD-CIE \cite{cui2023uncertainty}, a few-shot SSCL method, where the model trains on 100 classes under full supervision in the first task, and each subsequent task trains 10 classes with SSL settings, including 5 labeled images per class.

\noindent\textbf{Metrics.} 
We adopt the average incremental top-1 accuracy as our primary evaluation metric: $A_{Avg}=\frac{1}{T}\sum_{t=1}^t A_t$,    where $A_t$ is incremental accuracy for task $t$, defined as $A_t=\frac{1}{t}\sum_{t=1}^t a_{t,i}$ and $a_{t,i}$ is the accuracy on test set of $i^{th}$ task after learning the $t^{th}$ task. Additionally, we report the final model accuracy $A_{Last}$ in the last task as a reference.

\noindent\textbf{Implementation Details.} We use ResNet-32 \cite{he2016deep} as the feature extractor $F$ for the CIFAR-10 and CIFAR-100 and ResNet-18 \cite{he2016deep} for the ImageNet-100 and CUB. For unlabeled data, we apply the data augmentation approach from FixMatch \cite{sohn2020fixmatch}. The projection head $P$ is a simple linear layer outputting $512$-dimension features (\ie, $d=512$), which is consistent with the dimension of the generated ETF vectors across all datasets. All loss weights $\lambda_{\mathtt{uns}}$, $\lambda_{\mathtt{cl}}$, $\lambda_{\mathtt{fsr}}$ and $\lambda_{\mathtt{cud}}$ are set to 1, and all temperature parametes $\beta$, $\gamma$ and $\xi$ are set to 0.1. Following \cite{sohn2020fixmatch}, the confidence threshold $\tau$ is set to 0.95. We utilize the SGD optimizer with a momentum of 0.9 and the weight decay is set to $10^{-5}$. For SSCL tasks, we use a memory buffer of size 5120, set the batch size to 64, and use a learning rate of 0.03. We train for 200 epochs with a 10-epoch warm-up followed by a cosine scheduler to reduce the learning rate. For few-shot SSCL, we follow the settings of UaD-CIE \cite{cui2023uncertainty} for the training parameters (see Sec. \ref{apx:axpermental_setting} of  supplementary for details).

\begin{table}[t]
\centering
\caption{Comparisons on 10-task ImageNet-100. We re-run NNCSL  under our setting for a direct comparison. See supplementary Sec. \ref{app:np} for results under the original NNCSL protocol.
}
\scriptsize
\setlength{\tabcolsep}{3.0mm}{
\begin{tabular}{c|cc|cc}
\toprule
\multirow{2}{*}{\textbf{Method}} & \multicolumn{2}{c|}{\textbf{ImageNet100-13}} & \multicolumn{2}{c}{\textbf{ImageNet100-100}} \\
 & \textbf{Avg} & \textbf{Last} & \textbf{Avg} & \textbf{Last} \\
\midrule
iCaRL \cite{rebuffi2017icarl} & 19.89 & 12.88 & 30.78 & 16.68 \\
NNCSL \cite{kang2023soft} & 42.19 & 33.64 & 56.78 & 53.84 \\
\midrule
iCaRL\&Fix \cite{fan2024dynamic} & 26.37 & 15.58 & 37.49 & 21.02 \\
+ DSGD \cite{fan2024dynamic} & 28.35 & 19.14 & 50.53 & 32.10 \\
\rowcolor{GrayMedium} \textbf{+ \textcolor{uu}{\textbf{U}}\textcolor{ss}{\textbf{S}}\textcolor{pp}{\textbf{P}} (Ours)} & {43.91} & {35.40} & {56.84} & {50.36}\\
\midrule
DER\&Fix \cite{fan2024dynamic} & 35.40 & 29.22 & 61.96 & 52.91 \\
+ DSGD \cite{fan2024dynamic} & 35.73 & 31.53 & 62.27 & 52.82\\
\rowcolor{GrayMedium} \textbf{+ \textcolor{uu}{\textbf{U}}\textcolor{ss}{\textbf{S}}\textcolor{pp}{\textbf{P}} (Ours)} & \textbf{46.09} & \textbf{39.58} & \textbf{62.29} & \textbf{55.01}\\
\bottomrule
\end{tabular}
}
\label{tab:expr_result_imagenet100}
\end{table}
\noindent\textbf{Baselines.} For SSCL tasks, we primarily consider published SSCL methods, including the previous SOTA methods: DSGD \cite{fan2024dynamic} and NNCSL \cite{kang2023soft}, as well as classic methods like CCIC \cite{boschini2022continual} and ORDisCo \cite{wang2021ordisco}. 
{Our main competitors is DSGD \cite{fan2024dynamic}. For a fair comparison, we adopt the same base SSCL learners (\ie, iCaRL\&FixMatch and DER\&FixMatch).} Additionally, we consider converting traditional fully-supervised CL methods to SSCL setting for reference. Specifically, this adaptation involves using only labeled data during training and discarding all unlabeled data. We apply this approach to classic methods like iCaRL \cite{rebuffi2017icarl} and DER \cite{yan2021dynamically}. For few-shot SSCL task, we primarily compare against the previous SOTA method, UaD-CIE \cite{cui2023uncertainty}, and include several baselines used in its original paper.

\begin{table*}[t]
\centering
\caption{Performance comparisons on 11-task CUB. We  provide the test accuracy on different tasks and average accuracy. 
We replace UaD-CIE's uncertainty-based distillation  with CUD and incorporate FSR and CUP into its training pipeline.
For fairness, we use the official code of UaD-CIE to build USP on top of it and report the results based on our re-run of UaD-CIE.
}
\scriptsize
\setlength{\tabcolsep}{3.0mm}{
\begin{tabular}{ccccccccccccc}
\toprule
\multirow{2}{*}{\textbf{Method}} & \multicolumn{11}{c}{\textbf{Task ID}} & \multirow{2}{*}{\textbf{Avg}} \\
\cmidrule(r){2-12} 
 & \textbf{1} & \textbf{2} & \textbf{3} & \textbf{4} & \textbf{5} & \textbf{6} & \textbf{7} & \textbf{8} & \textbf{9} & \textbf{10} & \textbf{11} & \\
\midrule
SS-iCaRL \cite{cui2021semi} & 69.89 & 61.24 &55.81 &50.99 &48.18 &46.91 &43.99 &39.78 &37.50 &34.54 &31.33 & 47.29 \\
SS-NCM \cite{cui2021semi} & 69.89&61.91&55.51&51.71&49.68&46.11&42.19&39.03&37.50&34.54&31.33&47.33\\
SS-NCM-CNN \cite{cui2021semi} & 69.89 & 64.87 & 59.82 & 55.14 & 52.48 & 49.60 & 47.87 & 45.10 & 40.47 & 38.10 & 35.25 &  50.78 \\
Semi-SPPR \cite{zhu2021self} & 68.44 & 61.66 & 57.11 & 53.41 & 50.15 & 46.68 & 44.93 & 43.21 & 40.61 & 39.21 & 37.43 & 49.34\\
Semi-CEC \cite{zhang2021few}& 75.82 & 71.91 & 68.52 & 63.53 & 62.45 & 58.27 &57.62 & 55.81 &54.85 & 53.52 & 52.26 & 61.32 \\
Us-KD \cite{cui2022uncertainty} & 74.69 & 71.71 & 69.04 & 65.08 & 63.60 & 60.96 & 59.06 & 58.68 & 57.01 & 56.41 & 55.54 & 62.89\\
\midrule
UaD-CIE \cite{cui2023uncertainty}& 75.87 & 73.05	&69.50 & 65.61 &64.37&61.84&61.49&58.93&56.95& 56.21 & 55.66 & 64.33\\

\rowcolor{GrayMedium} + \textbf{\textcolor{uu}{\textbf{U}}\textcolor{ss}{\textbf{S}}\textcolor{pp}{\textbf{P}} (Ours)} & \textbf{78.21}&\textbf{74.48}&\textbf{71.81}&\textbf{68.16}&\textbf{67.58}&\textbf{64.77}&\textbf{62.87}&\textbf{62.27}&\textbf{59.97}&\textbf{60.10}&\textbf{60.55}&\textbf{66.43}\\
\bottomrule
\end{tabular}
}
\label{tab:expr_result_cub1}
\end{table*}
\begin{table*}[t]
\centering
\caption{Base and novel classes accuracy on 11-task CUB. Base classes refer to the classes used for fully supervised training on the first task, while novel classes refer to non-base classes trained on in subsequent tasks. 
}
\scriptsize
\setlength{\tabcolsep}{2.5mm}{
\begin{tabular}{cccccccccccccc}
\toprule
\multirow{2}{*}{\textbf{Method}} & \multirow{2}{*}{\textbf{Classes}} & \multicolumn{11}{c}{\textbf{Task ID}} & \multirow{2}{*}{\textbf{Avg}}\\
\cmidrule(r){3-13} 
 & & \textbf{1} & \textbf{2} & \textbf{3} & \textbf{4} & \textbf{5} & \textbf{6} & \textbf{7} & \textbf{8} & \textbf{9} & \textbf{10} & \textbf{11} & \\
\midrule
\multirow{2}{*}{SS-iCaRL \cite{cui2021semi} } & Base & 69.89 & 62.32 & 60.62 & 58.99 & 58.59 & 57.77 & 59.88 & 56.21 & 54.46 & 50.54 & 46.11 & 57.76\\
& Novel & - & 53.22 & 32.38 & 24.07 & 22.76 & 23.34 & 17.58 & 16.40 & 16.39 & 16.13 & 16.32 & 23.86 \\
\midrule
\multirow{2}{*}{SS-NCM-CNN \cite{cui2021semi}} & Base & 69.89 & 65.80 & 64.97 & 63.79 & 63.81 & 61.08 & 65.24 & 63.73 & 58.77 & 55.74 & 51.88 & 62.24\\
& Novel & - & 56.37 & 34.70 & 26.03 & 24.04 & 24.68 & 19.14 & 18.60 & 17.70 &17.79 &18.36 & 25.74 \\
\midrule
\multirow{2}{*}{UaD-CIE \cite{cui2023uncertainty}} & Base & 75.87&74.58&74.09&\textbf{73.46}&72.24&\textbf{71.68}&\textbf{71.33}&\textbf{70.50}&\textbf{70.15}&\textbf{69.27}&\textbf{69.13}&\textbf{72.03}\\
&Novel & - & 57.35&46.29&39.58&45.02&42.54&45.37&42.75&40.82&41.98&42.49 & 44.42 \\

\rowcolor{GrayMedium}  & Base & 
  \textbf{78.21} & \textbf{75.00} & \textbf{74.13} & 73.36 & \textbf{72.42} & 70.74 & 68.99 & 68.44 & 67.25 & 66.62 & 66.66 & 71.07 \\
\rowcolor{GrayMedium} \multirow{-2}{*}{+ \textbf{\textcolor{uu}{\textbf{U}}\textcolor{ss}{\textbf{S}}\textcolor{pp}{\textbf{P}}  (Ours)}} & Novel & - &\textbf{69.18} & \textbf{60.07} & \textbf{50.93} & \textbf{55.67} & \textbf{53.08} & \textbf{52.86} & \textbf{53.64} & \textbf{51.07} & \textbf{53.00} & \textbf{54.57} & \textbf{55.41} \\
\bottomrule
\end{tabular}
}
\label{tab:expr_result_cub2}
\end{table*}

\subsection{Main Results}

\textbf{CIFAR-10 and CIFAR-100.} We first report the results of different methods on CIFAR-10 and CIFAR-100 in \cref{tab:expr_result_cifar}. USP achieves the best performance across nearly all settings, showing even more pronounced advantages on the more challenging CIFAR-100. Both USP and DSGD are based on distillation, but on CIFAR-10/-100, USP achieves an approximately 1-6\% higher 
$A_{Avg}$ and 
$A_{Last}$ than DSGD using the same SSCL learner. These results indicate that USP is overall more robust than DSGD, specifically demonstrating a significant advantage in stability.

\noindent\textbf{ImageNet-100.}
The results on ImageNet-100 are shown in \cref{tab:expr_result_imagenet100}, where it can be seen that USP significantly outperforms DSGD in both average accuracy and final task accuracy. As with the results on CIFAR, USP demonstrates a more pronounced advantage when the amount of labeled data is smaller, indicating higher efficiency in utilizing unlabeled data. Notably, compared to DSGD, USP shows a smaller difference between the average and the final accuracy under the same task settings, suggesting that USP achieves more stable performance across different training task and is better at mitigating forgetting than DSGD.

\begin{figure}[t]
    \centering
    \begin{subfigure}[t]{0.49\linewidth} %
        \centering
        \includegraphics[width=\linewidth]{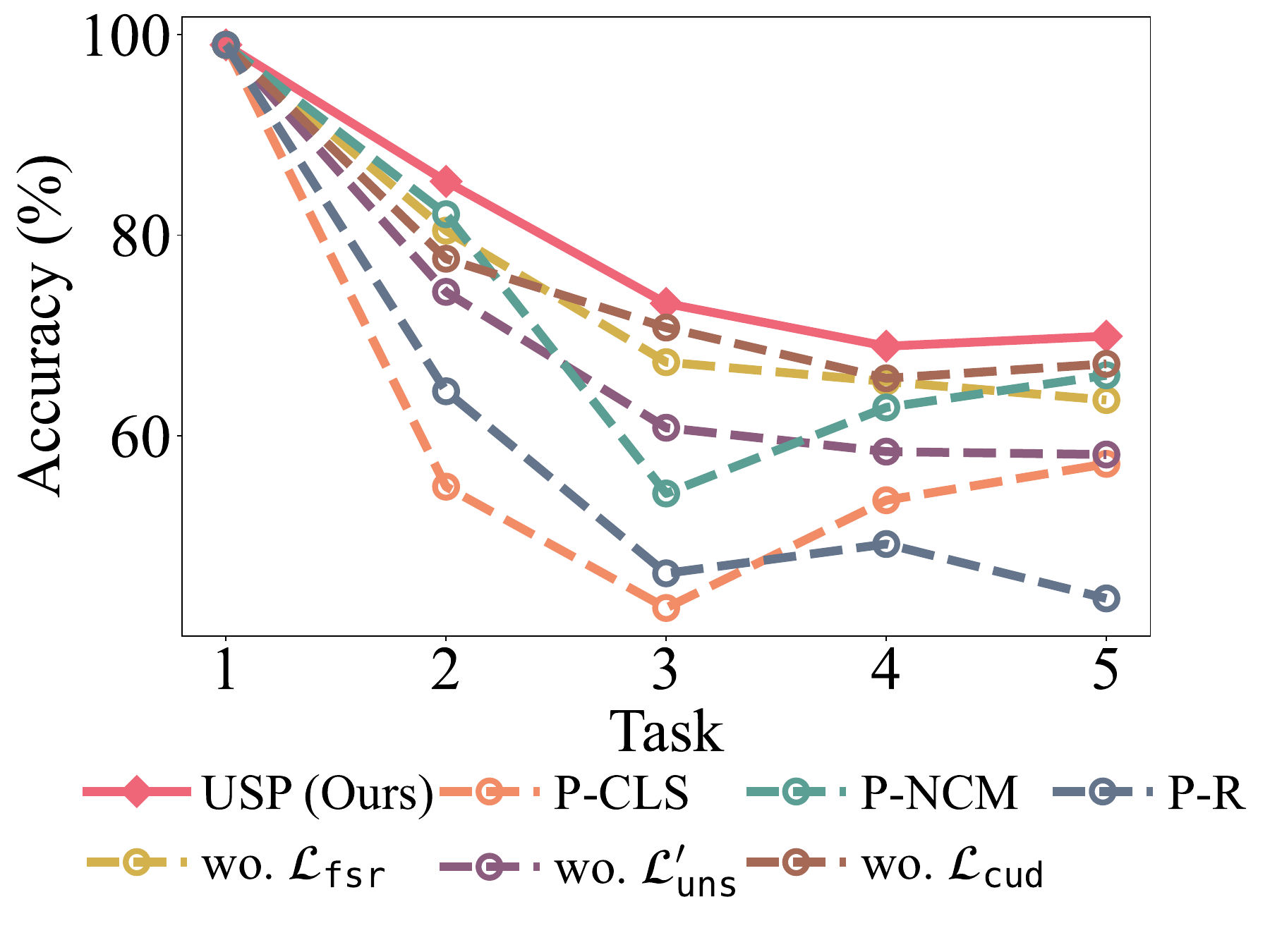} %
        \caption{Training phase} %
        \label{fig:abs_components1} %
    \end{subfigure}
    \begin{subfigure}[t]{0.49\linewidth} %
        \centering
        \includegraphics[width=\linewidth]{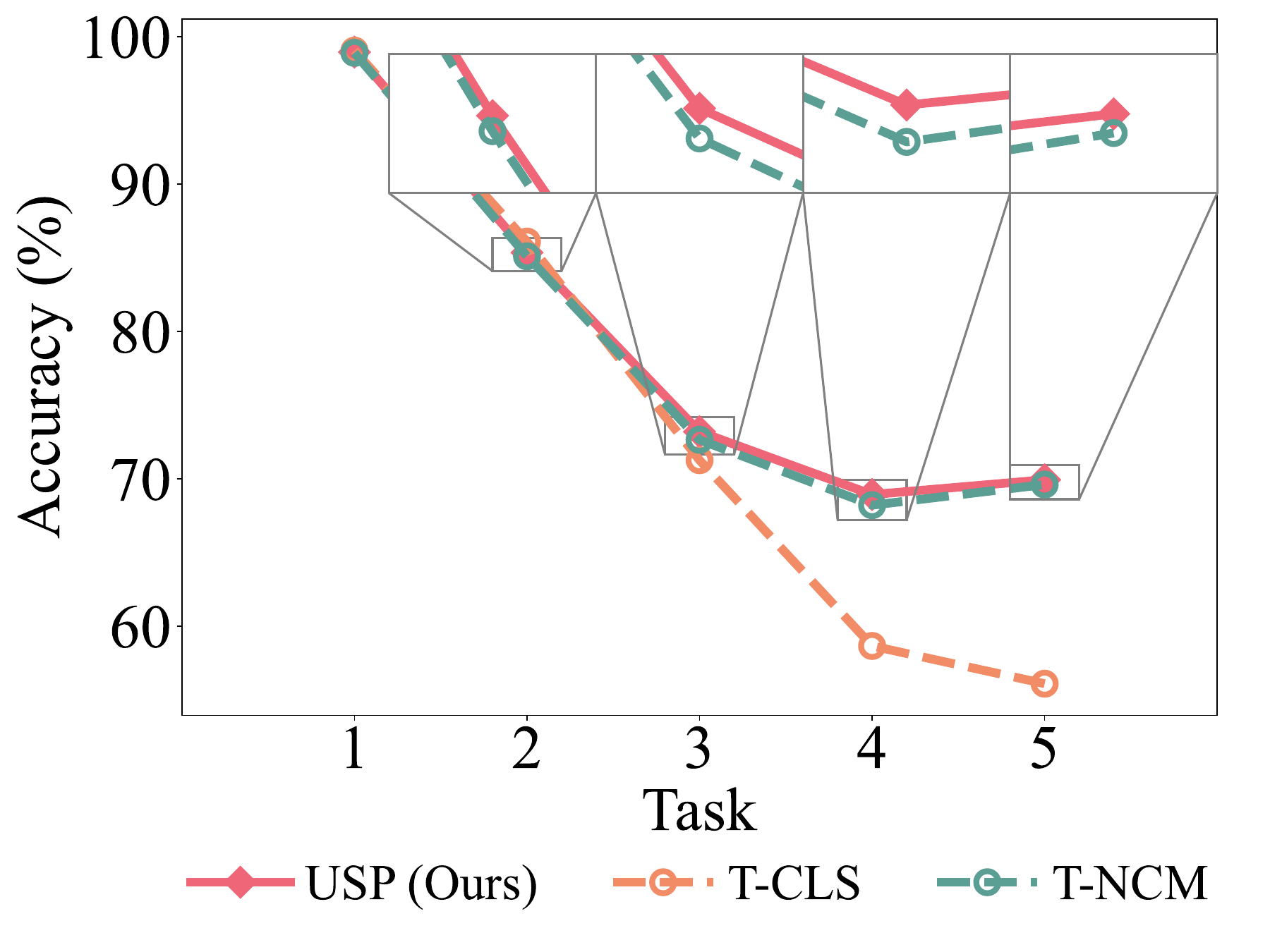} %
        \caption{Testing phase} %
        \label{fig:abs_components2} %
    \end{subfigure}

    \caption{Ablation studies on the main components of USP. The experiments are conducted on CIFAR-10 with 30 labels per class.}
    \label{fig:abs_components}
\end{figure}
\noindent\textbf{CUB.} The results on CUB are shown in \cref{tab:expr_result_cub1}. We conduct experiments on CUB following few-shot SSCL setting, comparing USP with the sota method UaD-CIE. USP can effectively improve the performance of UaD-CIE. To further illustrate USP’s effectiveness, we report base class and novel class accuracies across different training task, in \cref{tab:expr_result_cub2}. We observe that our method shows only a slight drop in base class accuracy compared to UaD-CIE (0.96\%), yet achieves a substantial improvement in novel class accuracy (10.99\%). UaD-CIE mitigates forgetting in base classes by applying uncertainty-based loss weighting to corresponding labeled samples. In contrast, USP leverages unlabeled samples and employs CUD distillation strategy to support old class learning. This effective distillation allows us to further enhance novel class learning through our DCP.%

In addition to the standard settings, more realistic SSCL settings are provided in Sec. \ref{apx:sscl} of supplementary.

\begin{figure*}[t]
    \centering
    \includegraphics[width=\textwidth]{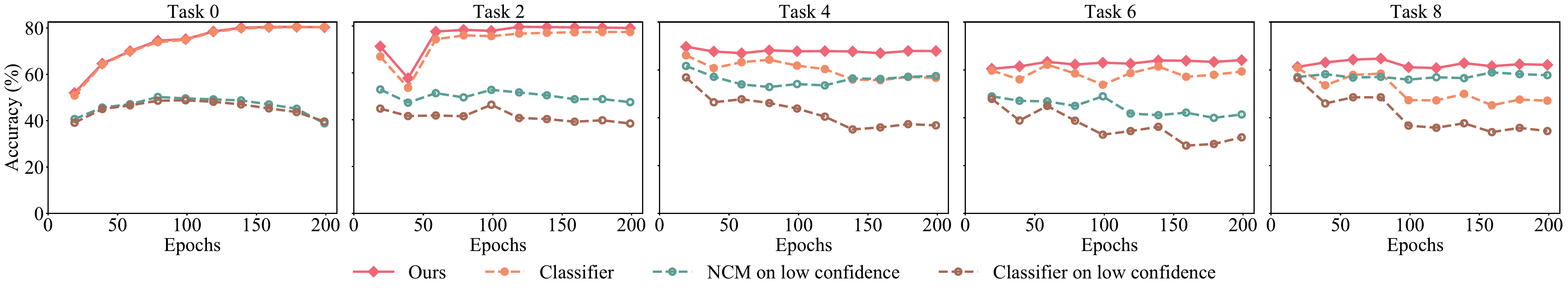}
    \caption{Pseudo-label accuracy of different pseudo-labeling strategies on CIFAR100-20. The y-axis represents the pseudo-label accuracy, and the x-axis represents the number of training epochs. We present the results for different training tasks.}
    \label{fig:abs_pseudo}
\end{figure*}
\subsection{Ablation Analysis and Discussions}
We perform extensive ablation studies on the components and training strategies of USP. 
By default, our experiments are all based on iCaRL\&Fix.
More ablations on distillation methods (Sec. \ref{app:dm}),  hyper-parameters (Sec. \ref{app:hyper}), backbones/pre-training (Sec. \ref{app:back}) and different memory buffer sizes (Sec. \ref{app:mbs}) can be found in the supplementary.

\noindent\textbf{Ablation on Method Components.}
We conduct ablation experiments on the main components of USP. With the basic SSCL learner training objective, USP incorporates the ETF-based feature space reservation (FSR) loss $\mathcal{L}_{\mathtt{fsr}}$, the unsupervised training loss $\mathcal{L}^\prime_{\mathtt{uns}}$ with our divide and conquer pseudo-labeling (DCP) strategy, and the class-mean-anchored unlabeled distillation (CUD) loss $\mathcal{L}_{\mathtt{cud}}$. To examine the effects of these modules, we ablate these losses, which is denoted as ``wo. $\mathcal{L}_{\mathtt{fsr}}$'', ``wo. $\mathcal{L}^\prime_{\mathtt{uns}}$'' and ``wo. $\mathcal{L}_{\mathtt{cud}}$'', respectively. As shown in \Cref{fig:abs_components1}, each component of USP contributes to improved model performance, with the DCP strategy having the most substantial impact on USP performance. To further investigate this, we conduct additional experiments by replacing DCP with alternative approaches: a classifier-based pseudo-labeling strategy (``P-CLS''), an NCM-based pseudo-labeling strategy (``P-NCM''), and a hybrid strategy where high-confidence samples use NCM pseudo-labels while low-confidence samples use classifier pseudo-labels (``P-R''). All these replacements lead to varying degrees of performance degradation, with P-R showing the most substantial decline, further underscoring the rationale behind our DCP. 
In \Cref{fig:abs_pseudo}, we further experiment to illustrate the reasons for its effectiveness.
Additionally, we perform an ablation study on DCP used in testing phase. \Cref{fig:abs_components2} presents results for classifier-only inference (``T-CLS'') and NCM-only inference (``T-NCM'') in isolation. Our DCP again achieves the best performance. While the improvement over NCM alone is not substantial, our strategy is a ``free lunch'' benefit.

\noindent\textbf{Effect of Pseudo-Labeling Strategy.} To demonstrate the effectiveness of DCP, we present the pseudo-label accuracy across different tasks on CIFAR100-20 in \Cref{fig:abs_pseudo}, comparing DCP with the traditional classifier pseudo-labeling strategy. Considering DCP fully utilizes unlabeled data, for a direct comparison of pseudo-label quality, we do not apply a confidence threshold to the classifier pseudo-labeling approach (see \cref{app:dcp} of supplementary for more comparison with thresholded pseudo-labeling). In the first task, our accuracy is comparable to the classifier's. However, as training progresses, DCP significantly outperforms the classifier, with an increasing advantage in later tasks.

\begin{figure}[t]
    \centering
    \includegraphics[width=\linewidth]{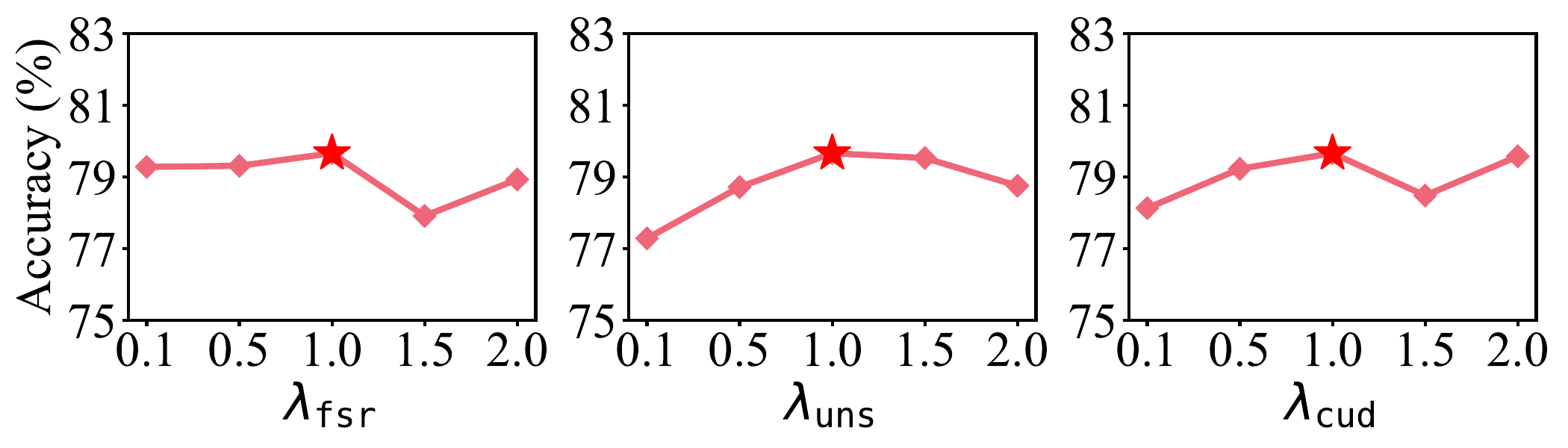}
    \caption{Average accuracy on CIFAR10-30 with different values of each loss weight (\ie, $\lambda_{\mathtt{fsr}}$, $\lambda_{\mathtt{uns}}$ and $\lambda_{\mathtt{cud}}$). 
    }
    \label{fig:abs_weight}
\end{figure}

During incremental training, the classifier faces stability-plasticity trade-offs, resulting in performance degradation. In contrast, the NCM classifier, leveraging feature similarity, is less affected by incremental training and achieves higher accuracy on low-confidence samples. DCP effectively combines the strengths of both methods, enhancing overall pseudo-label accuracy. To illustrate this, we report the pseudo-label accuracy of both approaches on low-confidence samples (confidence $< \tau$), where the NCM's higher accuracy affirms the rationale behind DCP.

\noindent\textbf{Loss Weights.} 
We traverse the values of $\{0.1, 0.5, 1, 1.5, 2\}$ to separately set the each weights (\ie, $\lambda_{\mathtt{fsr}}$, $\lambda_{\mathtt{uns}}$, and $\lambda_{\mathtt{cud}}$) while keeping the other two weights fixed at 1.0. 
We report the average incremental accuracy of the model on CIFAR10-30 under different weight coefficients, as shown in \Cref{fig:abs_weight}. The model achieves the best performance when the weights are set to 1.0, and we observe that the model's final performance fluctuates within 2\% across different weight coefficients, indicating that our method is not sensitive to the choice of weight coefficients. 

\section{Conclusion}
We propose a divide-and-conquer SSCL framework called USP, comprising three main components: (1) ETF-based feature space reservation (FSR) strategy for learning plasticity; (2) divide-and-conquer pseudo-labeling (DCP) approach for unlabeled learning; and (3) class-mean-anchored unlabeled distillation (CUD) for memory stability, which are designed to synergistically enhance the SSCL model. 
In future work, we plan to adapt our USP into more SSCL paradigms to further contribute to the community.

\section*{Acknowledgments}
Yue Duan, Taicai Chen and Yinghuan Shi are with the National Key Laboratory for Novel Software Technology and the National Institute of Healthcare Data Science, Nanjing University. Lei Qi is with the School of Computer Science and Engineering, Southeast University. This work is supported by NSFC Project (624B2063, 62222604, 62206052), China Postdoctoral Science Foundation (2024M750424), Fundamental Research Funds for the Central Universities (020214380120, 020214380128), State Key Laboratory Fund (ZZKT2024A14, ZZKT2025B05), Postdoctoral Fellowship Program of CPSF (GZC20240252), Jiangsu Funding Program for Excellent Postdoctoral Talent (2024ZB242) and Jiangsu Science and Technology Major Project (BG2024031).

{
    \small
    \bibliographystyle{ieeenat_fullname}
    \bibliography{main}
}

\clearpage
\appendix
\setcounter{page}{1}
\maketitlesupplementary

\definecolor{ss}{rgb}{1,0.4,0.4}
\definecolor{uu}{rgb}{0,0.6,0}
\definecolor{pp}{rgb}{0,0.4,0.8}

\section{More Implementation Details}
\label{apx:exper_detail}
\subsection{Exemplar Management}
\label{apx:exemplar_management}
We follow the exemplar management strategy of iCaRL \cite{rebuffi2017icarl}. Whenever the new classes are encountered, we adjust the exemplar set. All classes are treated equally, meaning that when $k$ classes have been observed so far and $M$ is the total number of storable samples, $m^t = \lceil M/k \rceil$  samples are allocated for each class at the $t$-th task. This ensures that the memory budget of $M$ samples is always fully utilized but never exceeded.  

Two routines are responsible for sample management: one for selecting samples for new classes and the other for reducing the size of the exemplar sets for previously classes. \Cref{alg:con_ex_set} outlines the sample selection process. Exemplars $e_1, \dots, e_m$ are selected and stored iteratively  until the target number $m$ is reached. At each iteration, a sample from the current training set is added to the exemplar set. The sample is chosen such that its feature vector brings the average feature vector of the exemplars closest to the average feature vector of the training samples. As a result, the exemplar ``set" is effectively a priority-ordered list, where the order of elements matters, and exemplars earlier in the list are more significant. The procedure for removing samples is specified in \Cref{alg:red_ex_set}, and it is particularly straightforward: to reduce the number of samples from any $m'$ to $m$, simply discard the samples $e_{m+1}, \dots, e_{m'}$, retaining only the exemplars $e_1, \dots, e_m$.

\subsection{Implementation Details For CUB}
\label{apx:axpermental_setting}
For CUB \cite{chaudhry2018efficient}, we follow the experimental setup and training pipeline of UaD-CIE \cite{cui2023uncertainty}. We use a base learning rate of 0.001 during the first task, which is divided by 10 after 80 and 120 epochs (out of a total of 160 epochs). For subsequent tasks, the learning rate is set to 0.0005, with a total of 60 supervised epochs. The training batch size is set to 32, and the testing batch size is set to 50. We use a memory buffer of size 2000, managed in accordance with iCaRL \cite{rebuffi2017icarl}. All loss weights $\lambda_{\mathtt{uns}}$, $\lambda_{\mathtt{cl}}$, $\lambda_{\mathtt{fsr}}$, and $\lambda_{\mathtt{cud}}$ are set to 1.0, and temperature parameters $\beta$, $\gamma$, and $\xi$ are set to 0.1.

\begin{algorithm}[t]
\caption{Constructing Exemplar Set}
\label{alg:con_ex_set}
\KwIn{Labeled dataset $D^{t,(i)}_l=\{x^{t,(i)}_{l,(1)}, \cdots, x^{t,(i)}_{l,(n)}\}$ of class $i$, target number of exemplars $m^t$, current feature extractor $F^t(\cdot)$.}
\KwOut{Exemplar set $E^{t,(i)}$}
$\mu_{D_l^{t,(i)}} = \frac{\sum_{x^{t,(i)}_{l} \in D_l^{t,(i)}}F(x^{t,(i)}_{l})}{|D_l^{t,(i)}|}$ \\
\For{$k = 1,\cdots, m^t$}{
$x_{e,(k)}^{t,(i)} = \underset{x_l^{t,(i)} \in D_l^{t,(i)}}{\text{argmin}}||\mu_{D_l^{t,(i)}}-\frac{1}{k}(F^t(x_l^{t,(i)})+\sum^{k-1}_{j=1}F^t(x_{e,(j)}^{t,(i)}))||$
}
$E^{t,(i)} = \{x_{e,(1)}^{t,(i)}, \cdots, x_{e,(m^t)}^{t,(i)}\}$
\end{algorithm}

\begin{algorithm}[t]
\caption{Reducing Exemplar Set}
\label{alg:red_ex_set}
\KwIn{Target number of exemplars $m^t$, exemplar set $E^{t-1,(i)}$ for class $i$}
\KwOut{Exemplar set $E^{t,(i)}$ for class $i$}
$E^{t,(i)} = \{x^{t-1,(i)}_{e,(1)}, \cdots, x^{t-1,(i)}_{e,(m^t)}\}$ \\
\end{algorithm}
\subsection{Building USP Based on DER}
\label{apx:axpintroduce_der}
DER \cite{yan2021dynamically} preserves the old network by parameter consolidation. At each incremental step, DER freezes previously learned representations and enhances them by adding new feature extractors, which introduce additional feature dimensions to the old representations. Additionally, DER introduces an auxiliary classifier $A(\cdot)$ to encourage the model to learn diverse and distinguishable features of new concepts. When constructing the USP based on DER, we follow DER’s dynamic network expansion during training while replacing $\mathcal{L}_{\mathtt{cl}}$ with DER’s corresponding training loss while keeping all other loss terms unchanged. Specifically, $\mathcal{L}_{\mathtt{cl}}$ is modified as:
\begin{equation}
\mathcal{L}_{\mathtt{cl}}(D^t \cup E^t, F^t)= \mathbb{E}_{x^t_l\sim D^t \cup E^t} \left[ H(\bar{p}^t_{x^t_l}, \bar{y}^t_l)\right] + \mathcal{L}_S(F^t),
\label{eq:cl_der}
\end{equation}
where, $\bar{p}^t_{x^t_l} = A^t(F^t(x^t_l))$ represents the prediction output of the auxiliary classifier $A^t(\cdot)$ introduced by DER. $A^t(\cdot)$ is a $(|\mathcal{Y}^t|+1)$-way classifier that treats all samples in the exemplar set $E^t$ as a single category. $\bar{y}_l^t$ represents the label, where $\bar{y}^t_l = y^t_l$ for $x_l^t \in D^t$ and $\bar{y}_l^t = |\mathcal{Y}^t|+1$ for $x^t_l \in E^t$. $\mathcal{L}_S(F^t)$ is the regularization loss computed based on the parameters of $F^t$ to prevent excessive model complexity. For detailed calculations, please refer to \cite{yan2021dynamically}.

\begin{table*}[t]
\centering
\caption{Performance comparisons on a 20-task continual learning benchmark under different data availability settings on ImageNet-100. We report both the original results of NNCSL \cite{kang2023soft} and the results of our own re-run (denoted as $^{\ast}$). In the original paper of NNCSL \cite{kang2023soft}, only the last accuracy is reported, without the average and task-level accuracy.
}
\resizebox{\textwidth}{!}{
\begin{tabular}{cccccccccccccccccccccccc}
\toprule
\multirow{2}{*}{\textbf{Labels}} & \multirow{2}{*}{\textbf{Method}} & \multicolumn{20}{c}{\textbf{Task ID}} & \multirow{2}{*}{\textbf{Avg}} \\
\cmidrule(r){3-22} 
 &  & \textbf{1} & \textbf{2} & \textbf{3} & \textbf{4} & \textbf{5} & \textbf{6} & \textbf{7} & \textbf{8} & \textbf{9} & \textbf{10} & \textbf{11} & \textbf{12} & \textbf{13} & \textbf{14} & \textbf{15} & \textbf{16} & \textbf{17} & \textbf{18} & \textbf{19} & \textbf{20} & \\
\midrule
\multirow{4}{*}{\textbf{1\%}} & NNCSL & - & - & - & - & - & - & - & - & - & - & - & - & - & - & - & - & - & - & - & 29.70 & -  \\
& NNCSL$^\ast$ &59.50&50.20&39.71&43.50 &	38.58 	&34.13 &	32.88& 	29.50 &	30.59 &	29.84 &	27.93 &	30.53&	31.09 &	30.37 &	30.22	&29.70	&29.62	&29.36	& 28.79 &28.98&34.25\\
& iCaRL\&Fix+\textcolor{uu}{\textbf{U}}\textcolor{ss}{\textbf{S}}\textcolor{pp}{\textbf{P}} & \textbf{64.80} & 50.80 & 52.93 & 49.50 & 44.80 & 39.67 & 34.97 & 34.55 & 32.49 & 31.48 & 29.27 & 33.13 & 33.48 & 34.57 & 34.05 & 33.12 & 32.87 & 31.29 & 30.40 & 28.64 & 37.84 \\
& DER\&Fix+\textcolor{uu}{\textbf{U}}\textcolor{ss}{\textbf{S}}\textcolor{pp}{\textbf{P}} &64.40&\textbf{55.00}&\textbf{53.33}& 	\textbf{51.10}& 	\textbf{47.12} &\textbf{43.13} &\textbf{ 41.60} &\textbf{ 41.00} & \textbf{38.58} & \textbf{37.08} & \textbf{35.64} &\textbf{ 38.10} &\textbf{37.78} & \textbf{36.91} & \textbf{36.53} & \textbf{34.20} & \textbf{33.48} & \textbf{33.09} & \textbf{33.64}& \textbf{32.78} & \textbf{40.00} \\

\midrule
\multirow{4}{*}{\textbf{5\%}} & NNCSL & - & - & - & - & - & - & - & - & - & - & - & - & - & - & - & - & - & - & - & 51.30 & - \\
& NNCSL$^\ast$ & 58.00 &	55.60 	&45.43 &	48.80 &	27.93 	&39.53 &	39.53 	&39.53 	&37.59 &	40.04 &	39.52 &	42.13	&42.31	&43.51	&43.16&	41.73	&39.40&	41.69	&42.43&	43.26  & 42.56  \\
& iCaRL\&Fix+\textcolor{uu}{\textbf{U}}\textcolor{ss}{\textbf{S}}\textcolor{pp}{\textbf{P}} & 73.60 	&62.40 &	68.00&	66.00& 	61.52 &	56.93 &	54.80 &	52.55 	&51.11 &	51.84 &	50.04 	&52.23	&51.85&	52.11&	52.40&	50.85	&49.81	&49.16	&49.05	&48.46&	54.56 \\
& DER\&Fix+\textcolor{uu}{\textbf{U}}\textcolor{ss}{\textbf{S}}\textcolor{pp}{\textbf{P}} &\textbf{76.00} &	\textbf{74.80} &	\textbf{72.00} 	&\textbf{72.00} &	\textbf{63.68} &	\textbf{60.20} 	&\textbf{58.63} 	&\textbf{57.10} 	&\textbf{54.93} &	\textbf{53.12} &	\textbf{53.20} 	&\textbf{55.20}	&\textbf{55.17}	&\textbf{55.63}	&\textbf{55.89}	&\textbf{54.70}&	\textbf{53.58}	&\textbf{53.53}&	\textbf{53.37}&	\textbf{53.62}	&\textbf{59.32} \\

\midrule
\multirow{4}{*}{\textbf{25\%}} &  NNCSL & - & - & - & - & - & - & - & - & - & - & - & - & - & - & - & - & - & - & - & \textbf{65.60} & - \\
& NNCSL$^\ast$ & 60.00 	&60.00 	&51.43 &	54.30 &	48.17 	&43.40 	&42.12 &	41.90 &	44.05 	&44.44 	&42.33& 	44.033	&45.53 &	46.14 &	45.78	&46.24&	43.53	&41.48	&41.67	&44.12&	46.53 \\

&iCaRL\&Fix+\textcolor{uu}{\textbf{U}}\textcolor{ss}{\textbf{S}}\textcolor{pp}{\textbf{P}} & 78.00 &	\textbf{77.00} &	79.73 &	78.50 &	71.60 &	\textbf{68.00} &	\textbf{65.09} &	\textbf{63.00} &	\textbf{60.13} &	\textbf{58.12} &	\textbf{57.83} &	\textbf{58.97} &	\textbf{59.82}&	58.17&	\textbf{59.07}&	\textbf{55.60}&	\textbf{55.48}&	54.49&	53.77	&53.78&	\textbf{63.31} \\
& DER\&Fix+\textcolor{uu}{\textbf{U}}\textcolor{ss}{\textbf{S}}\textcolor{pp}{\textbf{P}} & \textbf{80.40} 	&76.60 	&\textbf{79.87} 	&\textbf{79.20} 	&\textbf{71.76}& 	66.67 &	64.57 &	60.80 	&58.18 	&56.40& 	55.89 	&58.70	&57.66	&\textbf{58.57}&	55.39	&53.42&	51.04	&\textbf{54.69}&	\textbf{56.82}&	55.54 &	62.61 \\

\bottomrule
\end{tabular}
}
\label{tab:expr_result_20task_settings}
\end{table*}

\subsection{Neural Collapse and Equiangular Tight Frame}
\label{apx:etf_matrix}
Neural collapse refers to the phenomenon occurring at the late stage of training on balanced data (after the training error rate reaches 0). It reveals the geometric structure formed by the final layer features and the classifier, which can be defined as a simplex Equiangular Tight Frame (ETF), which refers to a matrix composed of $K$ vectors in $\mathbb{R}^d$, satisfying: 
\begin{equation}
E = \sqrt{\frac{K}{K-1}}U(I_K - \frac{1}{K}1_K1_K^T),    
\end{equation}
where $E=[e_1, \cdots, e_K]$. $U\in \mathbb{U}^{d\times K}$ allows a rotation and satisfies $U^\top U = I_K$ , $I_K$ is the identity matrix, and $1_K$ is an all-ones vector. All column vectors in $E$ satisfies:
\begin{equation}
e_{k_1}^\top e_{k_2} = \frac{K}{K-1}\delta_{k_1,k_2}-\frac{1}{K-1},\ \forall k_1,k_2 \in [1,K],    
\end{equation}
where $\delta_{k_1,k_2} = 1$ when $k_1 = k_2$, and 0 otherwise. All vectors have the same $L_2-$normalization and any pair of two different vectors has the same inner product of $-\frac{1}{K-1}$, which is the minimum possible  cosine similarity for $K$ equiangular vectors in $\mathbb{R}^d$. 

In our method, we use an simplex equiangular tight frame as the pre-defined class prototype features, with the sample features of each class aligned to it. More details about the neural collapse phenomenon can be found in \cite{yang2022neural}.

\section{Additional Experimental Results}
\label{apx:abla_exper}
Unless otherwise specified, DSGD \cite{fan2024dynamic} and USP both adopt iCaRL\&FixMatch \cite{fan2024dynamic} as the base SSCL learner.
\subsection{More SSCL Protocols}
\label{apx:sscl}
\subsubsection{NNCSL Protocol}
\label{app:np}
To ensure a comprehensive comparison with recent work, we conduct additional experiments to evaluate our method, USP, against NNCSL \cite{kang2023soft}. The original NNCSL protocol utilizes a different 20-task setting on ImageNet-100, which is distinct from our primary 10-task setup. To provide a fair comparison, we evaluate USP under NNCSL protocols. The results are presented in \cref{tab:expr_result_20task_settings}. The experiments show that USP consistently outperforms NNCSL across all settings, demonstrating the superior effectiveness and robustness of our approach.

\subsubsection{SSCL with Non-IID Distributions}
We consider two more realistic SSCL scenarios: (1) training with a long-tailed class distribution for each task (``\textit{imbalanced}''); (2) training with various data amounts across tasks (``\textit{inconsistent}''). Specifically, we conduct experiments on the 5-task CIFAR10-30. In the imbalanced setting, we set the number of labeled and unlabeled data for each class in each task to $\{30, 150\}$ and $\{600, 3000\}$. In the inconsistent setting, we set the training data sizes for the five tasks to $\{10000\to 250\to 125\to 5000\to 625\}$. The results are shown in \cref{tab:task_setting}. As can be seen,  our method demonstrates stronger robustness, with performance clearly outperforming the previous SOTA SSCL method.

\begin{table}[t]
\centering
\caption{Average and last accuracy on 5-task CIFAR10-30 with two more realistic SSCL settings. } 
\scriptsize
\begin{tabular}{c|cc|cc}
\toprule
\multirow{2}{*}{\textbf{Method}}  & \multicolumn{2}{c|}{\textbf{Imbalanced}} & \multicolumn{2}{c}{\textbf{Inconsistent}} \\
 & \textbf{Avg} & \textbf{Last} & \textbf{Avg} & \textbf{Last} \\
\midrule
DSGD & 62.42 & 62.96 & 57.58 & {59.92} \\
\textcolor{uu}{\textbf{U}}\textcolor{ss}{\textbf{S}}\textcolor{pp}{\textbf{P}}  & \textbf{75.18} & \textbf{65.50} & \textbf{70.26} & \textbf{60.39}  \\
\bottomrule
\end{tabular}
\label{tab:task_setting}
\end{table}

\subsection{More Ablation Studies}
\begin{table}
\centering
\caption{Ablation experiments on whether uses low-confidence samples (``LCS'') on 5-task CIFAR10-30.}
\scriptsize
\begin{tabular}{c|cc}
\toprule
 &  \textbf{Avg} & \textbf{Last} \\
\midrule

wo. LCS & 68.34 & 61.01  \\
w. LCS & \textbf{81.43} & \textbf{73.65} \\
\bottomrule

\end{tabular}
\label{app:dcp}
\end{table}

\subsubsection{Utilization of Low-Confidence Unlabeled Data}
To present the contribution of DCP, we conduct the following ablation experiments on using the low-confidence unlabeled data: traditional classifier with thresholded pseudo-labeling v.s. our proposed DCP, which is shown in \cref{app:dcp}. This comparison demonstrates that reasonably learning from low-confidence samples, rather than simply discarding them to avoid potential errors, can indeed lead to tangible performance improvements.

\begin{table}[t]
\centering
\caption{Ablation studies on different distillations on 10-task CIFAR100-25.}
\scriptsize
\begin{tabular}{c|cc}
\toprule
{\textbf{Method}} &  \textbf{Avg} & \textbf{Last} \\
\midrule

logit & 53.91 & 37.97  \\

feature & 48.16 & 33.56  \\

\midrule

CUD & \textbf{54.36} & \textbf{38.25} \\
\bottomrule
\end{tabular}
\label{tab:abs_dis}
\end{table}

\subsubsection{More Distillation Methods} 
\label{app:dm}
We explore the use of existing distillation methods for distilling from unlabeled data, specifically logit distillation and feature distillation. In particular, we apply consistency regularization directly on the logits or features output by the models of the current task and the previous task for unlabeled data. These experiments are compared with our proposed CUD, which are shown in \cref{tab:abs_dis}. It is evident that our CUD outperforms both logit and feature distillation.

\subsubsection{Hyper-parameters}
\label{app:hyper}
\begin{figure}[t]
    \centering
    \includegraphics[width=0.49\linewidth]{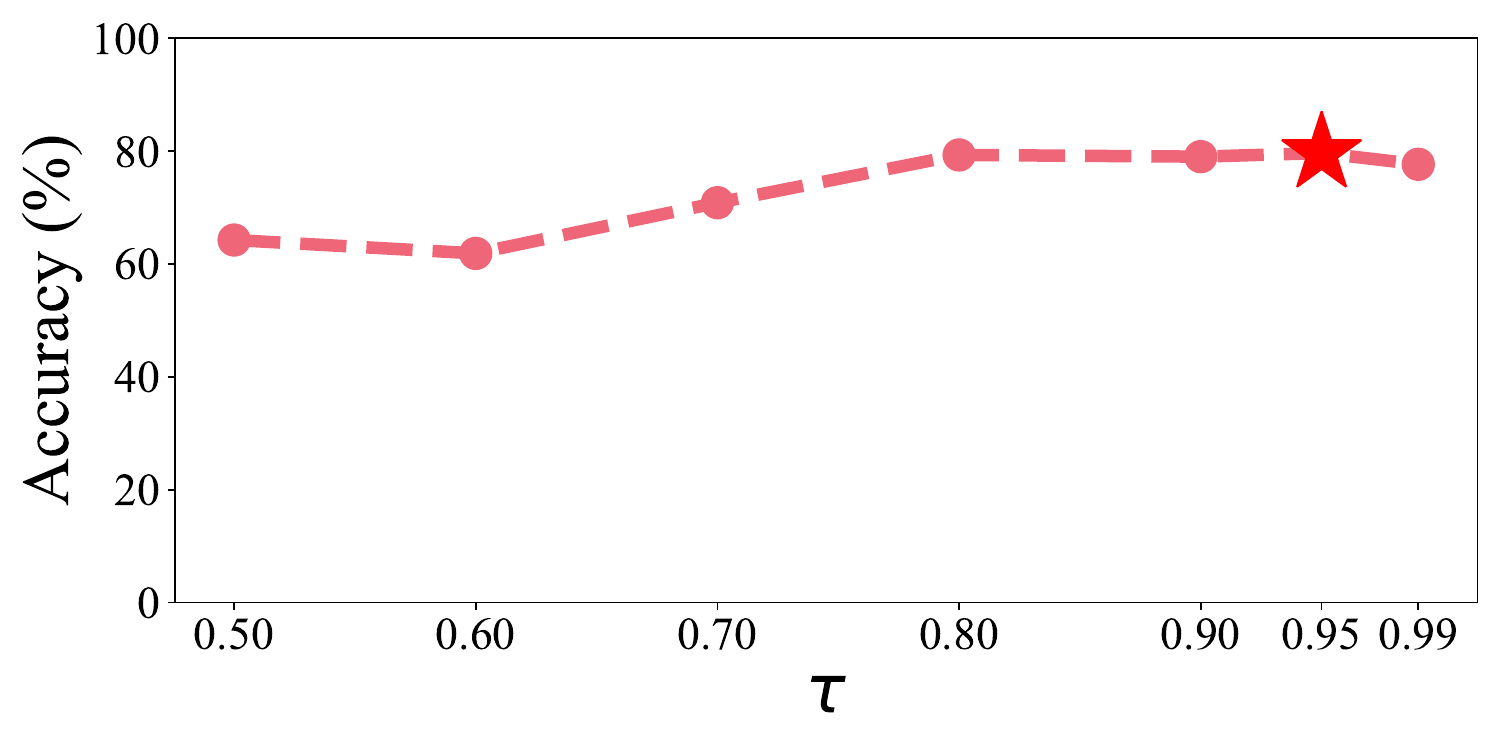}
    \includegraphics[width=0.49\linewidth]{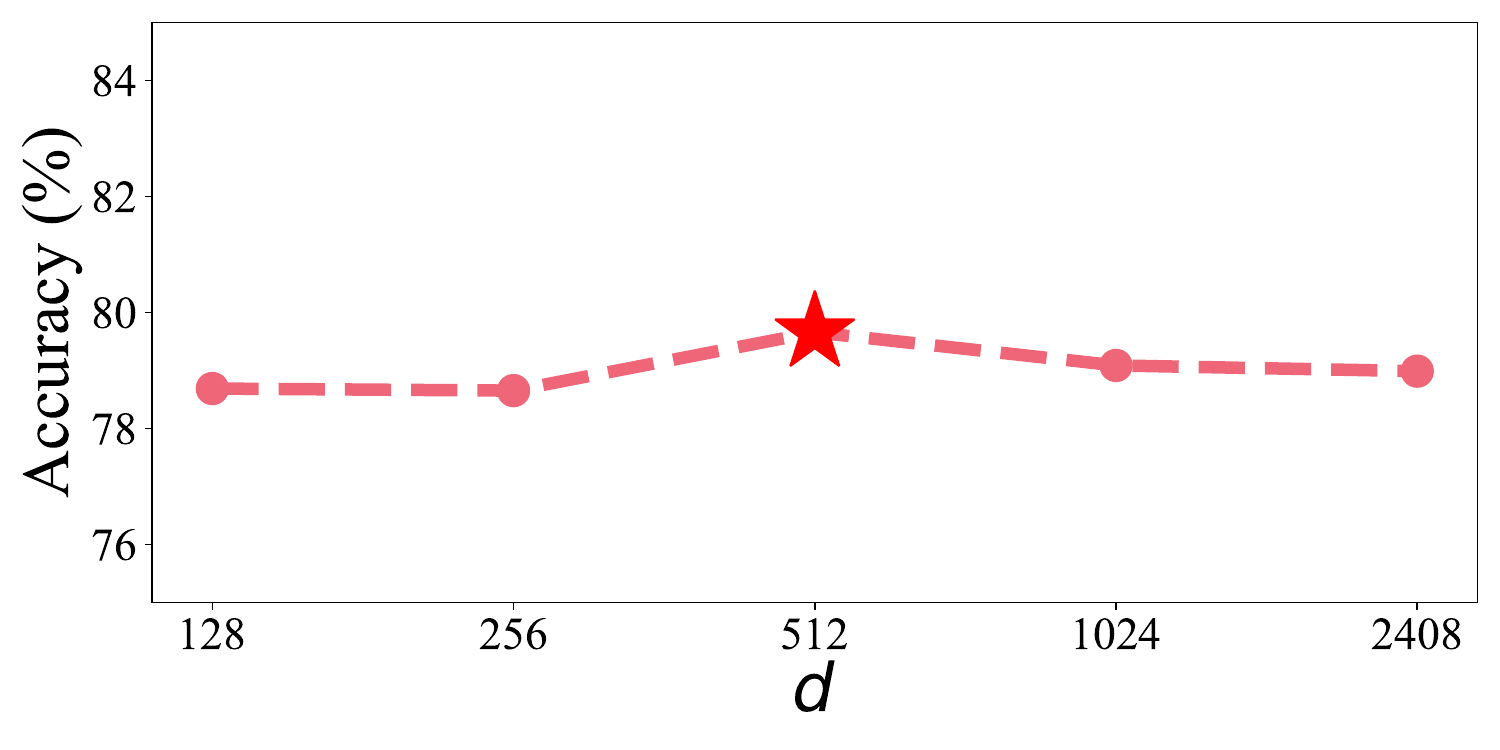}
    \caption{Average accuracy with various confidence thresholds and feature dimensions on 5-task CIFAR10-30.}
    \label{fig:ab}
\end{figure}

\noindent\textbf{Confidence Threshold and Feature Dimension.}
We conduct ablation studies on the confidence threshold $\tau$ and the feature dimension $d$. As \Cref{fig:ab} Shown, USP achieves the best performance with appropriately tuned default values. The threshold $\tau$ is set following standard practice in semi-supervised learning methods (\eg, FixMatch \cite{sohn2020fixmatch}), and the method demonstrates low sensitivity to variations in $d$.

\begin{table}
\scriptsize
\centering
\vspace{-1em}
\caption{Ablation studies on loss weights of $\mathcal{L}_{\mathtt{fsr}}$ on 5-task CIFAR10-30.}
\setlength{\tabcolsep}{1.7mm}{\begin{tabular}{cc|cc}
\toprule
$\lambda_{\mathtt{fsr}}^l$ & $\lambda_{\mathtt{fsr}}^u$ &  \textbf{Avg} & \textbf{Last} \\
\midrule
1.0 & 0.5 & 79.52 & 70.21  \\
1.0 & 1.0 & \textbf{81.63} & \textbf{73.65} \\
0.5 & 1.0 & 78.38 & 68.78  \\
\bottomrule
\end{tabular}} %
\label{tab:fsr}
\end{table}
\noindent\textbf{Loss Weights.} In our paper, the $\mathcal{L}_{\mathtt{fsr}}$ sums the labeled and unlabeled parts with the same weight. We further apply different loss weights to labeled and unlabeled data to investigate their impact on the performance of the method. We denote the loss weight for unlabeled data as $\lambda_{\mathtt{fsr}}^u$ 
and for labeled data as $\lambda_{\mathtt{fsr}}^l$, and conduct the corresponding ablation experiments. The experimental results are shown in \cref{tab:fsr}. The performance is best when the loss weights for labeled and unlabeled data are equal. Increasing or decreasing the relative weight of the unlabeled data leads to a performance drop, indicating that the pseudo-labels obtained through our divide-and-conquer labeling have high quality.

\subsubsection{More Backbones and Pre-Training Strategies}
\label{app:back}
In the main text, we follow the experimental setup of DSGD \cite{fan2024dynamic} and primarily use ResNet-32 and ResNet-18 without pre-training as the backbones for our method. To further investigate the impact of different backbones and pre-training strategies on the performance of our method, we use iCaRL\&Fix as the base SSCL learners and conduct ablation experiments. The experimental results are shown in \cref{tab:backbone}. We observe that using properly sized networks with appropriate pre-training leads to better USP performance. Simply using larger networks or advanced pre-training without proper adaptation does not guarantee improved SSCL performance (as found in \cite{lee2023pre}). Making USP more compatible with larger networks and diverse pre-training approaches remains our future work.
\begin{table}[t]
\centering
\caption{Ablation studies on different backbone architectures on the 5-task CIFAR10-30. Meanwhile, we adopt different pre-training strategies (CLIP \cite{radford2021learning} and DINO \cite{caron2021emerging}) on ResNet-50 to show the performance potential of our method.}
\scriptsize
\setlength{\tabcolsep}{0.8mm}{
\begin{tabular}{c|cc|cc|cc|cc|cc}
\toprule
\multirow{2}{*}{\textbf{Backbone}}  & \multicolumn{2}{c|}{\textbf{ResNet20}} &  \multicolumn{2}{c|}{\textbf{ResNet32}}  & \multicolumn{2}{c|}{\textbf{ResNet50}} & \multicolumn{2}{c|}{\textbf{CLIP}} & \multicolumn{2}{c}{\textbf{DINO}}\\
 & \textbf{Avg} & \textbf{Last} & \textbf{Avg} & \textbf{Last} & \textbf{Avg} & \textbf{Last} & \textbf{Avg }& \textbf{Last} & \textbf{Avg }& \textbf{Last}\\
\midrule

DSGD & 72.63 & {69.43} & 77.33 & \textbf{76.41} & 73.81 & {65.01} & 72.43 & 72.02 & 77.29 & 70.41\\

\textcolor{uu}{\textbf{U}}\textcolor{ss}{\textbf{S}}\textcolor{pp}{\textbf{P}} & \textbf{80.00} & \textbf{69.59} & \textbf{79.66} & {70.43} & \textbf{75.17} & \textbf{67.24} & \textbf{80.88} & \textbf{74.08} & \textbf{78.86} & \textbf{71.35}\\

\bottomrule
\end{tabular}
}
\label{tab:backbone}
\end{table}

\subsection{Discussions on Memory Buffer Size} 
\label{app:mbs}
By default, we follow the setup of iCaRL \cite{rebuffi2017icarl} and use a buffer size of 5120 to store a portion of the labeled data from each task as the exemplar set. To further investigate the impact of buffer size, we conduct additional ablation experiments, with the results presented in \cref{tab:abs_buffer}. As shown, a buffer size of 5120, which is the typical choice for most replay-based methods \cite{rebuffi2017icarl,boschini2022continual,kang2023soft}, achieves the best performance. Using a fixed-size exemplar buffer is a standard practice in continual learning \cite{rebuffi2017icarl,kang2023soft,fan2024dynamic}, as it reflects realistic memory constraints and enables fair comparisons with existing SSCL methods. While labeled data are indeed scarce in SSCL, the memory budget may still be insufficient to retain all labeled samples—particularly in settings with long task sequences (\ie, task ID $\rightarrow \infty$) or high supervision levels (\eg, CIFAR100-125 or ImageNet100-100, where the number of labeled samples reaches 12.5K and 10K, respectively, far exceeding the our default memory buffer size of 5120). In such scenarios, USP adopts an iCaRL-style exemplar buffer to strike a balance between memory efficiency and model performance. 

Although USP is designed under the realistic assumption of limited memory, our three key components—FSR, DCP, and CUD—are orthogonal to buffer size and remain effective even under larger or unlimited memory settings. Notably, DCP and CUD can also effectively leverage the unlabeled sample pool to address distribution shifts across tasks. To further verify the performance of USP under idealized conditions where the buffer is sufficiently large to retain all labeled samples, we conduct additional experiments on CIFAR100-125 and ImageNet100-100 with a buffer size of 20K. As shown in \cref{tab:bf20k}, USP continues to achieve strong performance in this setting, demonstrating the robustness and generality of our approach.
\begin{table}[t]
\centering
\caption{Ablation studies on memory buffer size of exemplar set $E^t$ on 5-task CIFAR10-30.}
\scriptsize
\setlength{\tabcolsep}{4.0mm}{
\begin{tabular}{c|cc|cc}
\toprule
\multirow{2}{*}{\textbf{Buffer Size}} & \multicolumn{2}{c|}{\textbf{CIFAR10-30}} & \multicolumn{2}{c}{\textbf{CIFAR10-150}} \\
 & \textbf{Avg} & \textbf{Last} & \textbf{Avg} & \textbf{Last} \\
\midrule

250 & 71.66 & 59.93  & 79.25 & 66.76  \\

500 & 73.21 & 61.75  & 80.71 & 72.48  \\

\midrule

5120 & \textbf{79.66} & \textbf{70.43} & \textbf{84.78} & \textbf{78.21}\\
\bottomrule
\end{tabular}
}
\label{tab:abs_buffer}
\end{table}

\begin{table}[t]
\centering
\caption{Comparisons with CL-based baselines (combine FixMatch \cite{sohn2020fixmatch} to exploit unlabeled data) using a larger buffer size 20K, which is enough to retain all labeled samples. }
\scriptsize
\setlength{\tabcolsep}{4.0mm}{
\begin{tabular}{c|cc|cc}
\toprule
\multirow{2}{*}{\textbf{Method}}  & \multicolumn{2}{c|}{\textbf{CIFAR100-125}}  & \multicolumn{2}{c}{\textbf{ImageNet100-100}}  \\
 & \textbf{Avg} & \textbf{Last} & \textbf{Avg} & \textbf{Last}  \\
\midrule
iCaRL\&Fix (20K)  & 62.07 & 46.56 & 40.40 & 26.91\\
+ \textcolor{uu}{\textbf{U}}\textcolor{ss}{\textbf{S}}\textcolor{pp}{\textbf{P}} (20K) & \textbf{68.65} & \textbf{55.17} & \textbf{56.91} & \textbf{51.73} \\
\midrule
DER\&Fix (20K)  & 68.75 & 54.83 & 62.02 & 53.46 \\
+ \textcolor{uu}{\textbf{U}}\textcolor{ss}{\textbf{S}}\textcolor{pp}{\textbf{P}} (20K) & \textbf{70.60} & \textbf{61.33} & \textbf{62.17} & \textbf{58.34} \\
\bottomrule
\end{tabular}
\label{tab:bf20k}
}
\end{table}

\end{document}